# Cross-lingual brain-language model alignment is robust but challenges hierarchical and computational accounts

***Ni Yang[1] *Rui He[1] Philipp Homan [2,3]**
**Iris Sommer[4] Davide Staub[5] Wolfram Hinzen[1,6]**
*These authors contributed equally.

1. *Grammar and Cognition Lab, Department of Translation & Language Sciences, Universitat Pompeu Fabra, Barcelona, Spain.*
2. *Department of Adult Psychiatry and Psychotherapy, University of Zurich, Zurich, Switzerland.*
3. *Neuroscience Center Zurich, University of Zurich and ETH Zurich, Zurich, Switzerland.*
4. *Center for Clinical Neuroscience and Cognition and Department of Psychiatry, University of Groningen, University Medical Center Groningen, Groningen, the Netherlands.*
5. *Scalable Scientific Machine Learning Lab, Imperial College London, Department of Earth Science and Engineering, London, United Kingdom.*
6. *Institut Català de Recerca i Estudis Avançats (ICREA), Barcelona, Spain.*

*Correspondence: ni.yang@upf.edu; rui.he@upf.edu*

**Abstract**

Brain-language model alignment is often interpreted as evidence that transformer models implement computations similar to those of the human brain. This assumes that neural predictivity reflects internal computational properties of large language models (LLMs), such as hierarchical contextual processing, predictive coding, or representational compression. An alternative possibility is that brain scores primarily reflect stable lexical-semantic correspondences shared by language models and the brain. Here we tested these competing interpretations using whole-brain encoding models across three typologically distinct languages: Mandarin, English, and French, during naturalistic story listening. Across all three languages, transformer representations significantly predicted activity in a distributed network spanning classical language regions, transmodal cortical systems, and subcortical structures. These spatial patterns showed substantial cross-linguistic overlap and remained remarkably stable across transformer layers, providing little evidence that model depth systematically maps onto cortical processing hierarchies. Likewise, contextual transformer embeddings did not consistently outperform static lexical embeddings, despite providing some unique predictive variance. Finally, neither surprisal nor intrinsic dimensionality reproduced the layer-wise profile of brain scores, arguing against prediction and information compression as primary explanations for brain-LLM alignment. Together, these findings suggest that brain-LLM alignment is more robust across languages, transformer depth, and model architectures than previously appreciated, but less informative about shared computational mechanisms. Rather than supporting a direct correspondence between transformer and cortical hierarchies, our results are more consistent with neural predictivity reflecting stable representational structure preserved across model transformations than with a one-to-one correspondence between their underlying computations.

## INTRODUCTION

Large language models (LLMs) have emerged as one of the most successful computational models of human language. Across neuroimaging modalities, representations extracted from transformer-based models reliably predict neural activity during naturalistic language comprehension, often outperforming earlier distributional semantic models and hand-engineered linguistic features (Caucheteux et al., 2022; Caucheteux & King, 2022; Goldstein et al., 2022, 2024, 2025; Kumar et al., 2024; Lei et al., 2025; Schrimpf et al., 2021; Xiao et al., 2026). These findings have been widely interpreted as evidence that modern language models capture aspects of the computations performed by the human brain during language processing. Yet the interpretation of these brain scores remains surprisingly unclear. High neural predictivity demonstrates that language models contain representations informative about brain activity, but it does not by itself establish why they do so. In particular, it remains unresolved whether brain–LLM alignment reflects shared computational mechanisms, such as contextual processing, hierarchical representations, predictive coding, or information compression, or whether both systems instead encode overlapping statistical or semantic structure present in natural language. Distinguishing shared computation from shared representational structure is therefore central to interpreting what brain–language model alignment does, and does not, imply about the relationship between artificial and biological language systems.

One influential interpretation is that transformer models predict brain activity because their internal processing hierarchy resembles the hierarchical organization of language processing in the brain. Beyond their generally high overall encoding performance, transformer models show systematic variation in neural predictivity across model depth. Several studies report that progressively deeper layers provide better predictions of progressively higher-order cortical regions, a pattern that has been interpreted as evidence that transformer depth mirrors a cortical hierarchy of language processing. In fMRI, intermediate BERT layers have been reported to best predict activity in primary auditory cortex, whereas deeper layers preferentially predict regions such as the left middle temporal gyrus, angular gyrus, and frontal cortex (Kumar et al., 2024). Similarly, comparisons between the embedding layer and a deeper GPT-2 layer showed stronger higher-level effects in bilateral superior frontal cortex, posterior superior temporal gyrus, precuneus, and triangular and opercular parts of the inferior frontal gyrus, regions typically associated with higher-order language comprehension (Caucheteux et al., 2022). Low-dimensional analyses of transformer representations further support this view: Antonello et al. (2021) found that earlier model representations align more strongly with auditory cortex, whereas later representations correspond more closely to distributed semantic regions across bilateral temporal, parietal, and prefrontal cortex, including the angular gyrus and precuneus. More recent work has similarly linked layered model representations to temporally or functionally organized cortical processing hierarchies (Caucheteux et al., 2023; Goldstein et al., 2025; Heilbron et al., 2022; Mischler et al., 2024).

But what, precisely, does such a "hierarchy" refer to? Language processing extends beyond the classical peri-sylvian network (Fedorenko et al., 2011; Hickok & Poeppel, 2007) and is embedded within large-scale functional systems that themselves exhibit hierarchical organization (Margulies et al., 2016; Yeo et al., 2011). Both production and comprehension recruit distributed cortical networks spanning sensory, attentional, control, and transmodal systems (Braga et al., 2020; Li et al., 2023; Rossi et al., 2025). Naturalistic narrative comprehension further requires the integration of information over extended temporal windows: events must be linked across sentences, context must be maintained and updated, and semantic information must accumulate over time to form coherent situation models (Baldassano et al., 2017; Dominey, 2021). These operations recruit higher-order associative and default-mode systems that sit toward the transmodal end of the principal cortical gradient (Margulies et al., 2016). Evaluating whether transformer depth truly mirrors cortical hierarchy therefore requires more than identifying isolated regions whose preferred layer differs. It requires asking whether layer-wise neural predictivity exhibits a coherent large-scale organization across the whole brain. Although prior work demonstrates that different transformer layers can preferentially predict distinct regions, it remains unclear whether these correspondences reflect a common macroscale hierarchy shared by transformers and the brain or more local, analysis-dependent similarities.

An alternative interpretation is that neural predictivity depends substantially on representational information that is preserved across model depth rather than on a one-to-one correspondence between successive computational stages. Transformer layers retain substantial lexical-semantic information even as other properties of their internal representations change (Jawahar et al., 2019; Kumar et al., 2024). Consistent with this possibility, static semantic embeddings can account for considerable neural variance. For example, Goldstein et al. (2022) found that contextual GPT-2 embeddings predicted more electrodes overall than static embeddings and explained unique variance at a subset of sites, but GloVe still significantly predicted a large number of electrodes in auditory, motor, and classical language regions. Thus, contextual representations may provide additional information without rendering static lexical-semantic structure irrelevant. If brain scores are driven in substantial part by representational information preserved across computational transformations, neural predictivity may remain relatively stable across model layers and architectures even when the representations themselves are not identical. Under this view, contextual processing and hierarchical transformations may remain important to language modeling without necessarily explaining why model representations are predictive of neural activity.

Before evaluating these competing interpretations, it is necessary to establish the empirical scope of the phenomenon that they seek to explain. Distributed cortical predictivity from language-model representations is already well documented and should therefore be treated primarily as a validation point rather than as a novel finding. Naturalistic language comprehension recruits not only canonical temporal and frontal language regions but also transmodal systems involved in semantic integration, narrative processing, memory, and contextual updating (Aliko et al., 2023; Braga et al., 2020; Fernandino & Binder, 2024; Hong et al., 2024; Rossi et al., 2025; Simony et al., 2016). Less well characterized is whether comparable LLM-based predictivity systematically extends to subcortical structures. Connectivity, lesion, and neuroimaging studies implicate structures such as the hippocampus and basal ganglia in semantic processing, sequencing, contextual binding, memory integration, and extended discourse comprehension (Booth et al., 2007; Cocquyt et al., 2019; Duff & Brown-Schmidt, 2012; Kurczek et al., 2013). A whole-brain analysis that jointly examines cortical and subcortical systems therefore provides a broader characterization of where model-derived linguistic representations predict neural responses during continuous naturalistic discourse.

A second aspect of empirical robustness concerns language. A robust account of brain-model correspondence should generalize, at least in part, across typologically distinct languages when listeners comprehend the same underlying narrative despite substantial surface-level linguistic differences. Recent multilingual studies provide evidence for such cross-lingual convergence. Varda et al. (2025), for example, demonstrated transfer of encoding performance across languages, suggesting the existence of shared neural-model representational structure. Zada et al. (2025) likewise reported cross-language convergence in average encoding performance and showed that similarity between language-model embeddings across languages relates to neural encoding. These findings point toward a shared conceptual representational space, but they do not fully characterize which cortical and subcortical regions are consistently shared across languages and which remain language-specific. Cross-linguistic consistency therefore provides an important test of the generality of brain-LLM correspondence, even though such consistency alone cannot distinguish between shared computational mechanisms and shared representational structure.

A more discriminating test concerns how neural predictivity changes as model representations are transformed across depth. If transformer depth systematically mirrors a cortical processing hierarchy, successive layers should produce structured changes in the cortical regions they best predict. Conversely, if neural predictivity depends strongly on information preserved across layers, encoding performance may remain comparatively stable despite substantial changes in internal model representations. This distinction can be evaluated not only by comparing mean layer-wise performance but also by testing pairwise differences between layers, examining the stability of preferred-layer assignments, quantifying representational similarity across layers, and asking whether similar layer-wise patterns generalize beyond bidirectional masked-language models to autoregressive architectures. Such analyses provide a more direct test of whether limited layer differentiation reflects genuine stability in neural predictivity rather than a peculiarity of a specific architecture or layer-selection procedure.

Meanwhile, contextualization provides a complementary mechanistic test. Transformer architectures are frequently assumed to outperform earlier embedding models because attention enables the construction of context-sensitive representations (Vaswani et al., 2023). Several studies support this interpretation. Schrimpf et al. (2021) reported higher normalized neural predictability for transformer-based models than for GloVe across fMRI and ECoG datasets. Antonello et al. (2021) similarly observed higher mean voxel-wise prediction correlations for transformer embeddings than for GloVe, and Kumar et al. (2024) found that BERT embeddings significantly outperformed GloVe in regions including Heschl's gyrus, posterior temporal cortex, angular gyrus, ventromedial prefrontal cortex, dorsomedial prefrontal cortex, and premotor cortex. However, a cleaner test compares not only non-contextual models but also the non-contextual input embedding from the same model and quantifies their shared and unique predictive contributions within a joint encoding framework. If contextualization is a major source of neural predictivity, contextual representations should explain substantial unique variance beyond the non-contextual representation and should show a systematic advantage in regions expected to support higher-order contextual processing.

Finally, even if neural predictivity varies across model depth, the computational quantity responsible for this variation remains uncertain. Two candidate processes have received particular attention: prediction and representational compression. Surprisal provides an index of predictive uncertainty and is motivated by the hypothesis that both biological and artificial language systems engage in predictive processing and error minimization during comprehension (Caucheteux et al., 2023; Goldstein et al., 2022; Heilbron et al., 2022; Schrimpf et al., 2021). However, recent work suggests that surprisal alone may not account for neural predictivity as robustly as previously assumed (Antonello & Huth, 2024; Cheng et al., 2026; Cheng & Antonello, 2024). Intrinsic dimensionality (ID), by contrast, characterizes the effective dimensionality of a representational space and has been used to quantify changes in abstraction, compression, and restructuring across transformer depth. Several studies report that ID profiles exhibit systematic trajectories across layers and, in some models, approximate the layer-wise profile of brain alignment (Baroni et al., 2026; Cheng et al., 2025, 2026; Cheng & Antonello, 2024). At the same time, ID varies across architectures, model scales, and training stages (Cheng et al., 2025; Cheng & Antonello, 2024), leaving open how general such relationships are. If surprisal or intrinsic dimensionality explains the observed organization of brain scores, their layer-wise trajectories should systematically correspond to the trajectory of neural predictivity. Conversely, a dissociation would provide limited support for these specific quantities as primary explanations of the observed layer-wise encoding profile, without implying that predictive processing or representational compression are absent from either system.

Here, we use multilingual whole-brain encoding models during naturalistic story comprehension to ask what aspects of brain-LLM correspondence are empirically robust and what computational interpretations this robustness can support. We first characterize the spatial distribution of neural predictivity across cortical and subcortical systems and its consistency across Mandarin, English, and French. We then test whether this correspondence systematically reorganizes across transformer depth, whether preferred-layer assignments are stable, and whether comparable patterns generalize to autoregressive language models. Next, we assess contextualization more directly by comparing contextual representations with the non-contextual BERT input embedding and by partitioning their shared and unique cross-validated predictive variance. Finally, we examine whether layer-wise surprisal and intrinsic dimensionality account for the observed brain-score profile. This progression separates the empirical phenomenon of neural predictivity from increasingly specific explanations of its computational origin, providing a principled framework for interpreting what brain scores reveal, and what they do not establish about the relationship between language models and the human brain.

## MATERIALS AND METHODS

### Dataset

The multilingual fMRI dataset was obtained from the publicly available "The Le Petit Prince: A multilingual fMRI corpus using ecological stimuli" at https://openneuro.org/datasets/ds003643/versions/2.0.5 (J. Li et al., 2022). The participants included 49 healthy young native English speakers (30 females; mean age = 21.3, *SD* = 3.6), 35 healthy, right-handed native Chinese speakers (15 females; mean age = 19.3, *SD* = 1.6), and 28 native French speakers (15 females, mean age=24.4, *SD*=4.6). The English audiobook, a 94-minute translation by David Wilkinson, was narrated by Karen Savage. Meanwhile, the Chinese version, approximately 99 minutes long, was read by a professional female broadcaster hired by the experimenter. The French audiobook is 97 minutes long, read by Nadine Eckert-Boulet and published by the now-defunct Omilia Languages Ltd. Both English and Chinese participants underwent a single fMRI scanning session, divided into nine runs of roughly 10 minutes each. During each run, participants passively listened to one segment of the audiobook and answered four comprehension questions afterward. The entire scanning session lasted approximately 2.5 hours.

The English and Chinese fMRI data were acquired using a 3T MRI GE Discovery MR750 scanner (Discovery MR750, GE Healthcare, Milwaukee, WI) with a 32-channel head coil. French MRI images were acquired with a 3T Siemens Magnetom Prisma Fit 230 scanner. Functional scans were acquired using a multi-echo planar imaging (ME-EPI) sequence with online reconstruction (TR=2000 ms; English and Chinese: TEs=12.8, 27.5, 43 ms; French: TEs=10, 25, 38 ms; FA=77◦; matrix size=72 x 72; FOV=240.0 mm x 240.0 mm; 2112x image acceleration; English and Chinese: 33 axial slices; French: 34 axial slices; voxel size=3.75 x 3.75 x 3.8 mm).

### fMRI preprocessing

Raw anatomical and functional MRI data from all three datasets were independently preprocessed using the same pipeline implemented in fMRIPrep 21.0.1 (Esteban et al., 2019), based on Nipype 1.6.1 (Gorgolewski et al., 2011). T1-weighted images underwent intensity non-uniformity correction, skull stripping, and tissue segmentation into cerebrospinal fluid (CSF), white matter (WM), and gray matter. Cortical surfaces were reconstructed using FreeSurfer, and anatomical images were nonlinearly normalized to the MNI152NLin2009cAsym template using ANTs. For the functional data, head motion was estimated relative to the BOLD reference using FSL MCFLIRT, yielding six rigid-body motion parameters, and slice-timing correction was performed using AFNI 3dTshift. The BOLD reference was co-registered to the participant's T1w reference using FreeSurfer boundary-based registration with six degrees of freedom. The BOLD time series were subsequently resampled into MNI152NLin2009cAsym standard space using the transformations estimated during motion correction, anatomical co-registration, and spatial normalization, with volumetric resampling performed using ANTs and Lanczos interpolation. After fMRIPrep preprocessing, nuisance regression was performed using the pipeline described by Wang et al. (2024), regressing out the effect of head motion, WM, and CSF, and adding discrete cosine basis regressors to handle low-frequency signal drifts. The resulting functional signals were subsequently averaged within the 400-parcel cortical parcellation (Schaefer et al., 2018) and 16 subcortical parcels from Melbourne Subcortex Atlas (S1) parcellation (Tian et al., 2020), yielding 416 regional time series for subsequent encoding analyses (Figure 1B).

We selected Schaefer-400 a priori because it provides an intermediate functional-connectivity-based cortical resolution and assigns every cortical parcel to one of the seven canonical Yeo functional networks. This organization allowed both parcel-level inference and network-level summaries while retaining sufficient signal within individual parcels. To evaluate whether differences in data quality could confound the cortical-subcortical comparisons, we additionally quantified temporal signal-to-noise ratio (tSNR) from the pre-normalization ROI time series. Within-atlas parcel-stability analyses and assessment of temporal signal-to-noise ratio are reported separately in Supplementary Information.

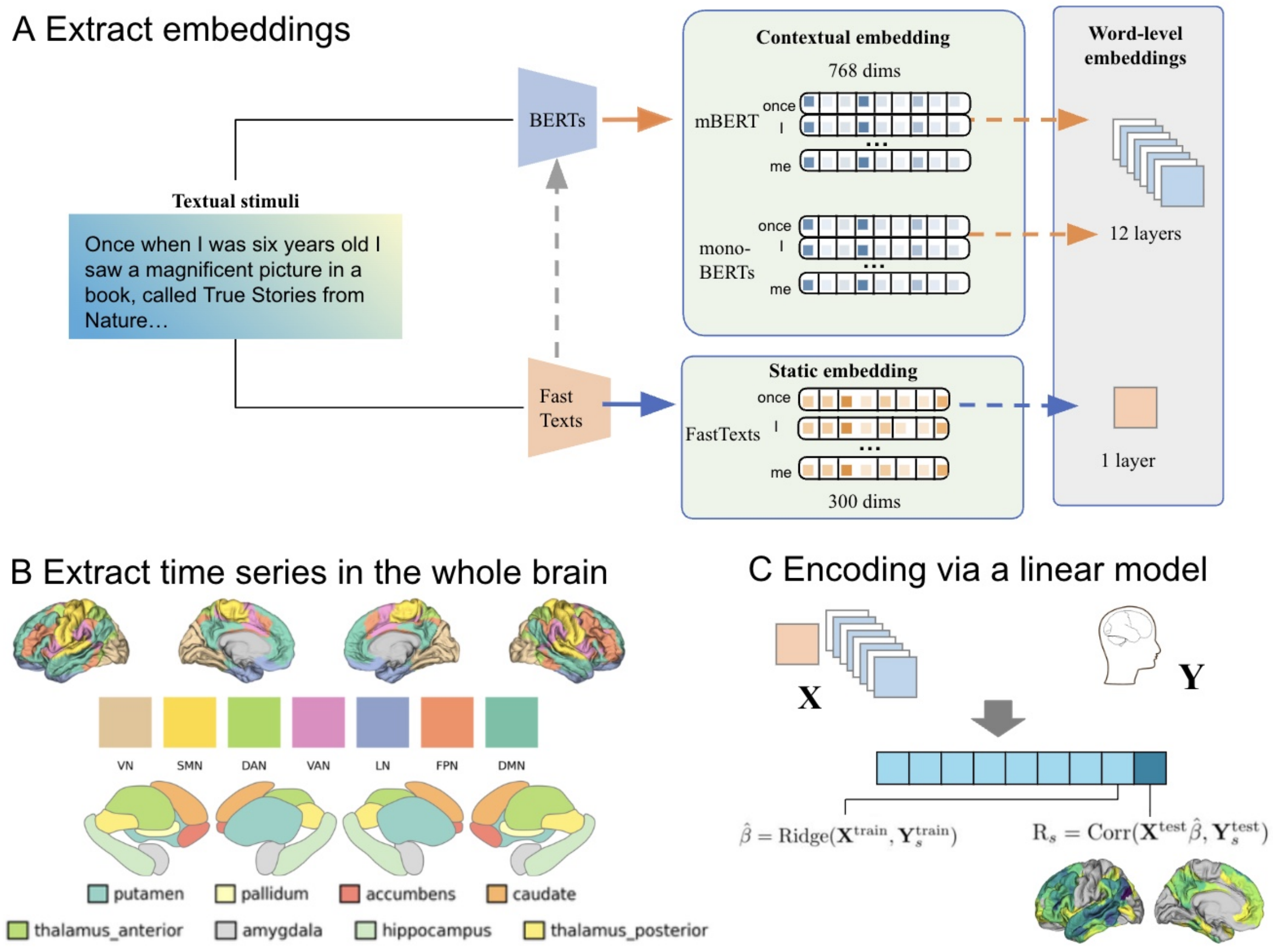

**Figure 1**: Encoding workflow. (A) The stimuli used for the storytelling were divided into different utterances. In addition to static semantic features (FastText), we extracted the contextual features from multilingual BERT (mBERT) and monolingual BERTs (monoBERT) for Mandarin, English, and French. (B) After preprocessing the fMRIs of subjects listening to fMRIs, time series in each ROI were extracted. The Yeo 7 functional networks: visual (VN), sensorimotor (SMN), dorsal attention (DAN), ventral attention (VAN), limbic (LN), frontoparietal (FPN), and default mode (DMN), and 16 subcortical regions from the Melbourne S1 atlas are shown above. (C) The encoding model was estimated from a training subset of each story using ridge regression (*X*) and evaluated on a left-one-out test segment using cross-validation (*Y*).

**Retrieving model embeddings**

The general workflow is shown in Figure 1. Primarily, 7 different models (multilingual BERT, 3 mono-lingual BERT models, and 3 FastText models, one for each language) were employed. Pre-trained FastText word vector models (cc.en.300.bin, cc.zh.300bin, and cc.fr.300bin) served as models for non-contextual word embeddings. For contextual embeddings, we use off-the-shelf BERT variants, including mBERT-base-cased, BERT-base-uncased, Chinese-BERT-wwm (HFL), and camembert-base (almanach) from the HuggingFace library (Wolf et al., 2020). The FastText model represents words as continuous vectors based on their overall usage patterns in the training corpus. It provides static embeddings that capture the general, conceptual meaning of words, which is effective for encoding broad and generic semantic relationships but cannot differentiate between different senses of a word within specific contexts. BERT variants use a deep transformer model that processes text bi-directionally, generating contextualized embeddings that capture the dynamic and specific meaning of words by considering their surrounding vocabularies within the attention window. Additionally, as a robustness check against architectural differences, we also extracted representations from GPT-2 (openai-community/gpt2) and Mistral (mistralai/Mistral-7B-v0.1) for the English stimulus. Both are causal autoregressive models, such that the representation at each word is conditioned only on the current and preceding context.

In retrieving word embeddings (Figure 1A), the FastText model processes each utterance as a sequence of individual words, generating a static 300-dimensional word embedding for each word. To proceed with our context-sensitive BERT-variant models, the annotated text was first segmented into lists of utterances, enabling the entire utterance to serve as the context for the given word. It is worth noting that previous studies have found that the context window affects both brain score (Mischler et al., 2024) and predicted areas in the brain (Pasquiou et al., 2023). While many studies use the maximum available context length (Antonello et al., 2021; Caucheteux et al., 2023; Varda et al., 2025), others have found that shorter windows, typically around 10 to 20 words (Toneva & Wehbe, 2019; Yu et al., 2024) or 50 words (Raugel et al., 2025), yield better alignment between LLMs and neural data. In our case, given the findings from former studies and in order to preserve the completeness of a sentence to ensure a meaningful context, as in its training objectives, we use the entire current sentence (15 words on average) as the context window. This choice minimizes ambiguity for the model and allows for more precise embedding generation.

The layer-wise contextual embeddings were obtained by passing the tokenized input through BERT, which returns hidden states from all layers. Each tensor has the shape of (number of layers, number of words, 768), which comprises a list of layer embeddings with 768 dimensions. These hidden states are extracted, including the initial encoding embedding layer as a built-in non-contextual representation from the model (referred to as layer 0 in the follow-up sections). Each layer's embeddings are then aligned with the original tokens by averaging the embeddings of the corresponding sub-tokens. To verify that the limited differentiation in neural encoding scores across mBERT layers was not attributable to redundant hidden-state extraction, we also quantified pairwise representational similarity across the embedding layer and 12 Transformer blocks using linear centered kernel alignment (CKA).

**Representation prediction and evaluations**

The relationship between language model representations and regional BOLD responses was estimated using ridge-regression encoding models. Ridge regression is a type of linear regression that includes a regularization term in the loss function that helps to prevent overfitting by shrinking the coefficients. Its efficiency in handling multicollinearity and managing high-dimensional data makes it suitable for encoding the embedding vectors and fMRI time series (Nunez-Elizalde et al., 2019). Following fMRI preprocessing and confound regression, the BOLD signal was parcellated into 400 cortical ROIs from the Schaefer-400 atlas and 16 subcortical ROIs from the Melbourne S1 atlas, yielding one time series per ROI and run. All the encoding analyses were performed directly at the ROI level rather than at the voxel or surface-vertex level. For each language-model layer, word-level feature vectors were first aligned to the fMRI sampling grid by convolving feature time courses with the canonical haemodynamic response function (HRF) according to the annotated word onsets using a package developed from a previous study (Pasquiou et al., 2022).

As a robustness analysis, we additionally replaced the canonical HRF with a finite impulse response model for a 12th layer in English mBERT, which does not impose a fixed haemodynamic response shape (Antonello et al., 2021; Caucheteux et al., 2023; Caucheteux & King, 2022; Huth et al., 2016). In this method, each annotated word onset was assigned to an fMRI volume by taking the floor of the onset time divided by the TR. When multiple words occurred within the same TR, their feature vectors were averaged. The resulting TR-level feature matrix was expanded using five causal FIR delays corresponding to 0, 2, 4, 6, and 8 s. The five delays were entered into the ridge regression simultaneously, and encoding performance was evaluated for the complete five-delay model on held-out runs. The HRF and FIR analyses otherwise used the same fMRI data, model representations, preprocessing, cross-validation, regression, and evaluation procedures.

Then, feature matrices and ROI time series were standardized within run, and the resulting run-wise design matrices were retained as separate temporal segments. Encoding models were eventually evaluated using an outer nine-fold leave-one-run-out cross-validation procedure. In each fold, all time points from eight complete runs were concatenated to form the training set. Because the remaining complete run was held out for evaluation, individual TRs were never randomly shuffled or interleaved across the training and test sets. For each fold, a ridge regression model was fit exclusively to the eight training runs. Regularization strength was selected from 10 logarithmically spaced values ranging from $10^{-3}$to $10^{8}$using RidgeCV, with regularization selected separately for each ROI. The fitted model was subsequently used to predict the 416 ROI time series of the held-out run. Encoding performance, which we refer to as the brain score, was defined for each subject, ROI, and fold as the Pearson correlation between the predicted and observed BOLD time series in that held-out run,

$$r_{s,j,k} = corr(\hat{y}_{s,j,k}, y_{s,j,k})$$

where ($s$), ($j$), and ($k$) denote subject, ROI, and held-out run, respectively. The nine held-out-run correlations were averaged to obtain one subject-level brain score for each ROI and model representation. Of note, in order to avoid the overfitting issue raised by Feghhi and colleagues (2024), data splits respected the original temporal order of the fMRI recordings. Model performance was evaluated using a leave-one-run-out cross-validation scheme with nine folds, strictly corresponding to the nine fMRI runs. In each fold, one run was held out for testing while the remaining runs were used for training (Figure 1C), so this held-out run was not supplied to the regression estimator during model fitting or regularization selection and was only used once. For each subject and ROI, encoding performance was quantified as the correlation between predicted and observed fMRI responses in the held-out data and was averaged across cross-validation folds to yield a single subject-level encoding score per ROI.

To assess whether the results were sensitive to the choice of regression model, in addition to the separate L0- and L12-based encoding models, we fitted a joint model using banded ridge regression using the English data. Banded ridge regression extends standard ridge regression by assigning separate regularization weights to different feature spaces, allowing the model to adjust the relative contributions of the L0 and L12 representations rather than treating their concatenation as a single feature set. The model performance was also quantified as the Pearson correlation between predicted and observed BOLD responses in held-out runs as comparable metrics, and $R^2$for unique variance between L0 and L12 predictions. Unique variance was calculated from matched cross-validated predictive estimates as $U_{L0} = R^2_{joint} - R^2_{L12}$ and $U_{L12} = R^2_{joint} - R^2_{L0}$. Group-level significance was the same as the ones mentioned before.

**Significant ROI predictions and preferred layer across languages**

To identify which ROIs have significant predictions at the group level, subject-wise encoding scores were aggregated into a subjects-by-ROIs matrix. For each ROI, we performed a one-sample one-sided t-test across subjects to assess whether encoding performance was significantly greater than zero. To correct for multiple comparisons across the 416 ROIs, $p$ values were adjusted using the Benjamini-Hochberg false discovery rate (FDR) procedure with a threshold of $q < 0.05$. ROIs surviving FDR correction were retained as significant, and non-significant ROIs were masked.

Overlap in significant cortical regions across languages was visualized by projecting ROI-level results onto the fsaverage5 cortical surface (Fischl et al., 1999) using the Schaefer-400 parcellation and Melbourne S1 subcortical atlas. For each mBERT layer, thresholded ROI maps from Chinese, French, and English were binarized based on the presence or absence of significant encoding effects. Cortical and subcortical regions were categorized as language-specific (Chinese-only, French-only, and English-only) or shared across all three languages. These categorical maps were converted from ROI space to surface plot using label-based mapping and rendered on the left and right hemispheres with BrainSpace (Vos de Wael et al., 2020).

Next, to investigate whether layer preference aligns with the language processing hierarchy in the brain, we derived parcel-wise preferred layers from group-level encoding performance maps for each language. At each layer, subject-level scores were averaged across participants to yield a single group-level performance vector per layer. For each cortical parcel, the preferred layer was defined as the model layer yielding the maximum mean encoding performance across layers, and the corresponding maximum performance value was retained, so every ROI was labeled with a model layer with the best performance. To assess the robustness of layer preference, we performed a subject-level bootstrap analysis separately for each language. We generated 5,000 bootstrap samples by resampling participants with replacement while preserving the original sample size. Within each bootstrap sample, encoding scores were averaged across the resampled participants for each layer and ROI. We then calculated, for each ROI and layer, the proportion of bootstrap samples in which that layer was selected as the preferred layer. Preferred-layer stability was defined as the maximum of these bootstrap selection probabilities across the 12 layers, such that higher values indicate that the same layer was consistently selected across participant resamples, whereas lower values indicate less stable layer preference. Additionally, in supplementary information, we also conducted a percentile-based sensitivity analysis to avoid the bias of only having max performance. For each language, participant-level encoding scores were first averaged within each of the 12 mBERT layers. Then, for each cortical parcel independently, we calculated the 80th, 85th, 90th, 95th, and 99th percentiles across the 12 layer-specific scores. Similar to max performance, layers with scores equal to or greater than the corresponding percentile threshold were retained, and the mean index of the retained layers was used as the parcel's percentile-based preferred-layer estimate.

### Layer difference and model comparison analysis

For each language and each region, we assessed differences in brain-model alignment across transformer layers using paired permutation tests across subjects. As mentioned earlier, brain scores were available for each subject, across all 12 layers and for all 416 ROIs. For every pair of layers, we computed within-subject differences in brain scores for each ROI and used the mean difference across subjects as the test statistic. Statistical significance was assessed using a paired sign-flip permutation test, in which the sign of each subject's difference was randomly inverted. This procedure corresponds to random within-subject exchanges of the two layer labels and preserves subject-level dependence while testing the null hypothesis that the two layers are exchangeable. Permutation-based $p$ values were computed independently for each ROI and corrected for multiple comparisons across ROIs using FDR with a threshold of $q$ =

0.05. To summarize layer-wise differences, we report the proportion of ROIs showing a significant between-layer difference after FDR correction. The same analysis was conducted on GPT2 and Mistral, which can be seen in the Supplementary Information.

For the mBERT vs. FastText performance comparison, we performed ROI-wise paired permutation tests across subjects. For each ROI, within-subject differences in encoding scores were computed between mBERT and FastText, and the mean difference across subjects was used as the test statistic. Statistical significance was assessed using a paired sign-flip permutation test, in which the sign of each subject's difference was randomly inverted, corresponding to random within-subject exchanges of model labels. This procedure preserves subject-level dependence while testing the null hypothesis that the two models are exchangeable. Two-sided permutation p-values were computed independently for each ROI and corrected for multiple comparisons across the 416 ROIs using the FDR ($q < 0.05$). As a robustness check complementing the ROI-wise permutation analyses, we additionally fit a pooled linear mixed-effects model to ROI-level encoding scores across subjects. For each language, model type (mBERT vs FastText) was treated as a fixed effect, while random intercepts were included for both subject and ROI to account for repeated measurements and variability across brain regions. The L0 and L12 of mBERT followed the same procedure (see in Supplementary Information).

**Compute layer-wise surprisal and intrinsic dimensionality**

As a measure of LLM's internal representation, we first computed layer-wise pseudo-log-likelihood (PLL) surprisal from multilingual BERT (mBERT; bert-base-multilingual-cased) using a masked-token scoring procedure (B. Li et al., 2021; Salazar et al., 2020). For each sentence, we first concatenated the pre-segmented word sequence into a string and tokenized it with the model's WordPiece tokenizer. We then iteratively masked each non-special token position, ran the masked sequence through mBERT, and extracted the hidden state at the masked position for each transformer layer. To obtain layer-specific token probabilities, we applied the model's masked-language-model (MLM) prediction head to each layer's hidden state and computed the log-probability of the token at that position. Surprisal was defined as the negative log-probability in nats. Repeating this for all token positions yields a PLL-based estimate of surprisal for each token conditioned on the remaining context. Because our linguistic annotations were at the word level, we aligned annotated words to WordPiece tokens using spacy-alignments and aggregated WordPiece surprisals by summing across the subword pieces corresponding to each word, producing a word × layer surprisal matrix (12 layers). Next, we aggregated the word-level surprisal at each layer and took the mean as the average layer surprisal.

To characterize the information compression process, we then acquired the ID of mBERT representations across layers by applying a nearest-neighbor–based estimator to token-level embedding spaces. For each language and each transformer layer, we extracted token embeddings from precomputed values from earlier. To control computational cost and sampling bias, we then randomly subsampled up to 8,000 tokens per language (without replacement, fixed random seed), yielding a set of high-dimensional vectors for each layer. Before ID estimation, embeddings were L2-normalized along the feature dimension, corresponding to a cosine-distance geometry. ID was then estimated using the two–nearest-neighbor maximum-likelihood estimator (2NN-ML) as implemented in the DADApy (Glielmo et al., 2022) package. We used this 2NN-ML estimator to estimate dimensionality from the ratio between the distances to the first and second nearest neighbors, under the assumption of locally uniform sampling on a low-dimensional manifold embedded in a higher-dimensional space. The resulting ID value reflects the effective number of degrees of freedom of the representation at each layer, independent of the ambient embedding dimension. This procedure yields a layer-wise ID curve for each language, allowing us to examine how representational complexity evolves across the depth of the model.

## RESULTS

### Distributed correspondence in both cortex and subcortex

We first identified the ROIs that the encoding models in each language significantly predict within each language. All three mBERT-based models exhibited significant predictive performance across a wide range of canonical language-processing regions, with substantial consistency across languages.

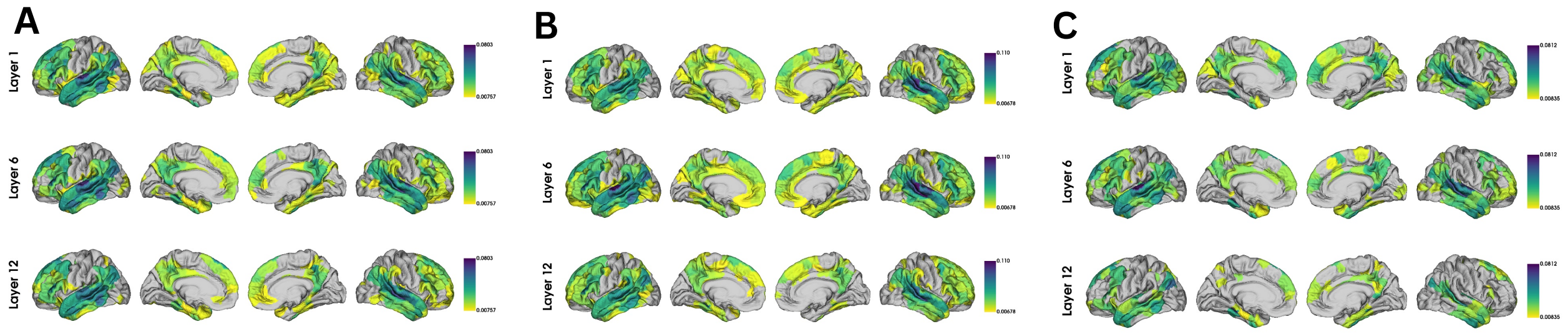

**Figure 2**: Significant cortical prediction effects across layers for English (A), Mandarin (B), and French (C) mBERT. Only regions with significant brain scores are colored. Statistical significance was assessed using one-sample t-tests of subject-level prediction scores against zero (FDR-corrected, $p < 0.05$). Layer indices correspond to the 12 Transformer encoder blocks of the model. The embedding layer was not included in the present analyses. Brain scores are reported as Pearson's $r$, with darker colors indicating higher values. Here are the three representative layers for early, intermediate, and later layers. Complete maps for all 12 mBERT layers are provided in the Supplementary Information, Figure S1-3.

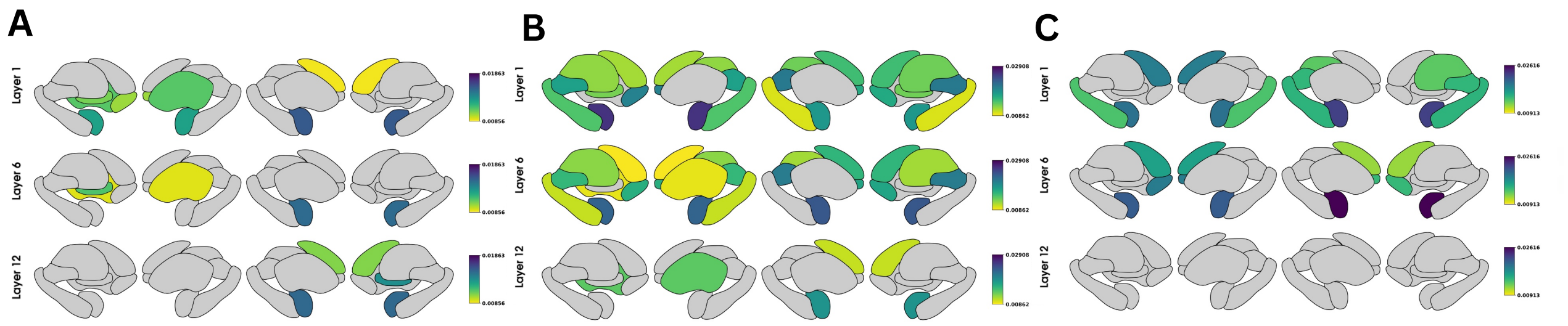

**Figure 3:** Significant subcortical prediction effects across layers for English (A), Mandarin (B), and French (C) mBERT. Only regions with significant brain scores are colored. Statistical significance was assessed using one-sample t-tests of subject-level prediction scores against zero (FDR-corrected, $p < 0.05$). Complete maps for all 12 mBERT layers are provided in the Supplementary Information, Figure S1-3.

Specifically, across layers (see Figure 2A), significant cortical prediction effects were observed in the posterior cingulate cortex (PCC), medial prefrontal cortex (mPFC), anterior cingulate cortex (ACC), and medial temporal lobe (MTL), as well as in regions canonically associated with higher-level language processing, including the inferior frontal gyrus (IFG), superior temporal gyrus (STG), middle temporal gyrus (MTG), temporal pole (TP), and inferior parietal lobe (IPL). The highest brain scores were consistently found in the primary auditory cortex. Collectively, the cortical results indicate stronger encoding performance in auditory regions, the default mode network, and higher-order language areas. In subcortical areas, caudate, putamen, and amygdala consistently yielded significantly higher scores than other subcortical areas (Figure 3). Similar spatial patterns were observed in French and Mandarin (Figure 2B-C and Figure 3B-C), supporting the view that language processing engages a distributed, whole-brain network.

### Overlap between predicted areas across languages

Next, to identify regions shared across all three languages, we examined the overlap of significantly predicted ROIs, as well as regions uniquely captured by each language. As shown in **Error! Reference source not found.**A, substantial overlap was observed across layers in both cortical and subcortical regions, specifically in pSTG, MTG, IFG, IPL, PCC, mPFC, and ACC in the cortex, as well as the amygdala in the subcortex. To further validate these patterns, we repeated the analysis using three language-specific models with stronger within-language performance (Figure S1, 2 and 3). The resulting shared areas closely mirrored the distributions in mBERT, indicating a robust language-agnostic spatial alignment. Despite the linguistic differences, the alignment reveals a shared language processing space across LLMs and brains.

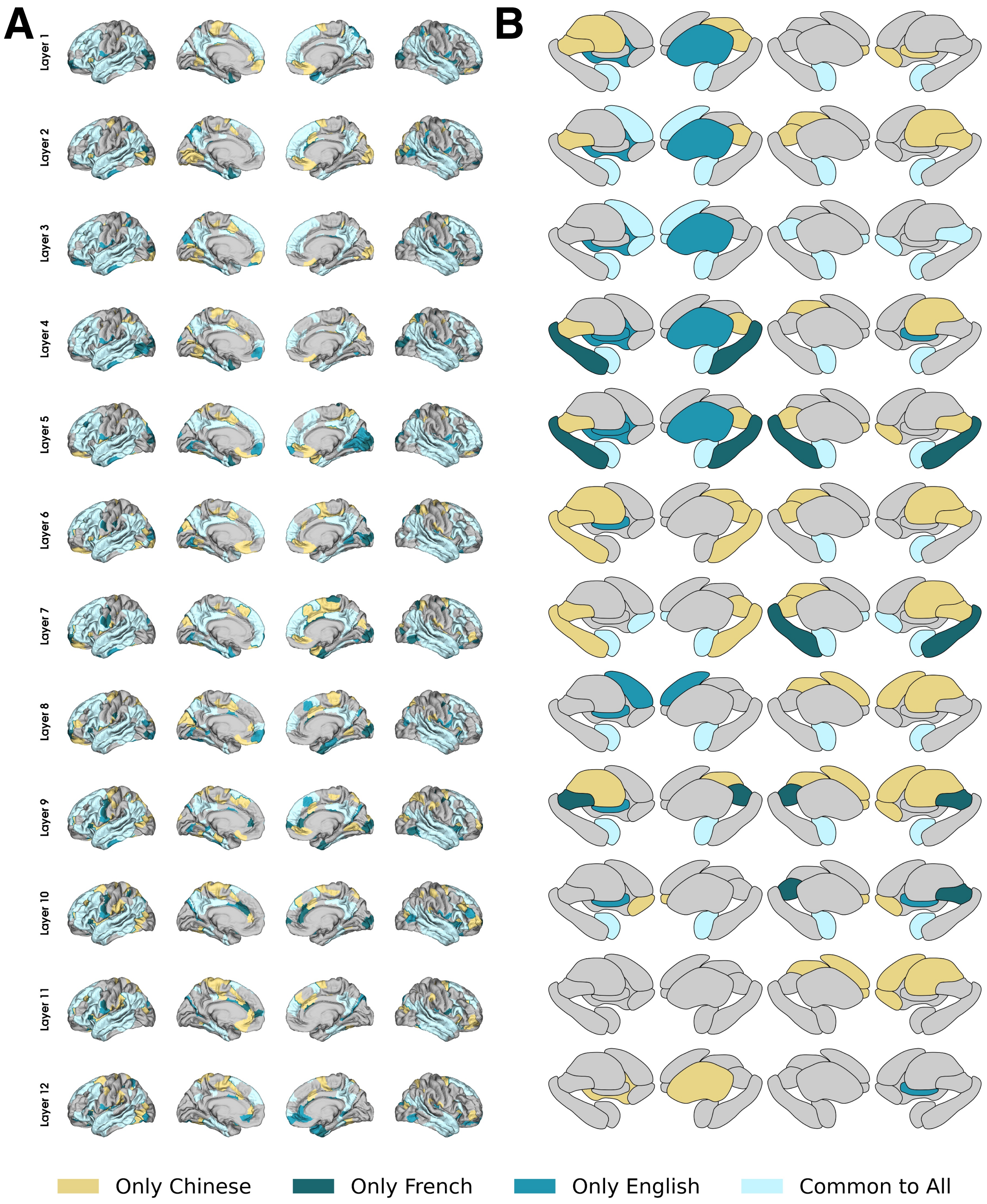

**Figure 4**: Overlapping predicted areas across languages in the cortex (A) and subcortex (B). Colored areas are ROIs with significant brain scores (FDR-corrected, $p < 0.05$). The overlapping predictions made by three languages are colored in light blue.

**No clear layer progression to reflect layer hierarchy**

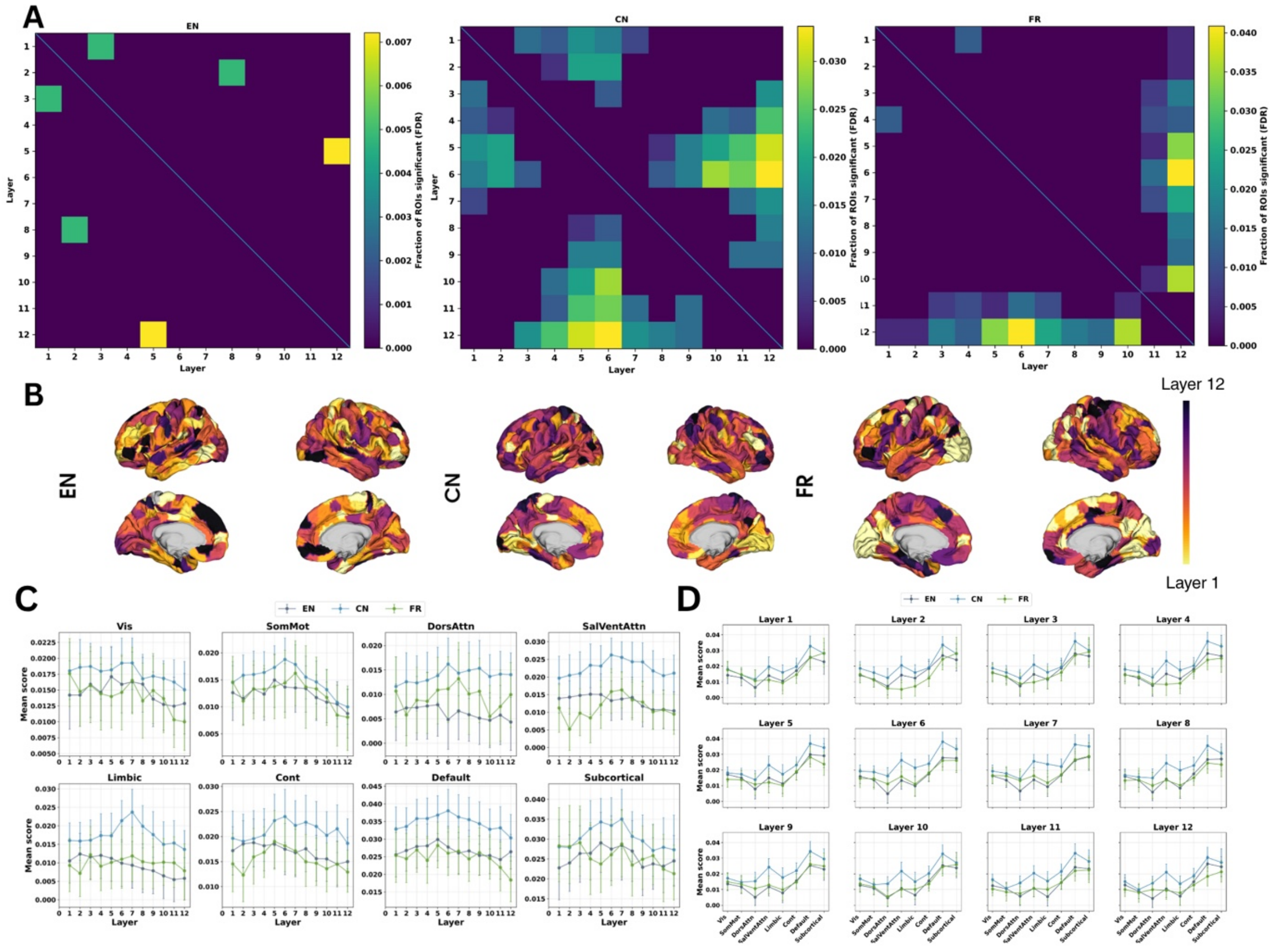

**Figure 5**: Layer-wise differentiation, preferred layers, and large-scale network profiles of mBERT brain scores across languages. (A) The heat map shows the fraction of ROIs (out of 416) whose brain scores differ significantly between layer *i* and layer *j*, based on a paired sign-flip permutation test across subjects, with FDR correction across ROIs ($q$ = 0.05). (B) The preferred layer at each ROI across three languages. The darker color indicates deeper model depth. Network-wise and layer-wise brain scores across languages. (C) Mean brain scores across mBERT layers within seven Yeo cortical networks and the Melbourne subcortical system: visual (Vis), somatomotor (SomMot), dorsal attention (DorsAttn), ventral attention (SalVentAttn), limbic (Limbic), control (Cont), default mode (Default) network, and the Melbourne subcortical system in 12 mBERT layers for Chinese (CN), English (EN), and French (FR). Error bars indicate 95% confidence intervals obtained from 10,000 participant-level bootstrap resamples with replacement. (D) Network profiles for each mBERT layer, showing mean brain scores across the seven cortical networks and the subcortical system for EN, CN, and FR. Error bars indicate 95% confidence intervals obtained from 10,000 participant-level bootstrap resamples with replacement..

To examine whether different transformer layers yield systematically different brain–model alignment patterns, we conducted pairwise layer comparisons using a subject-wise permutation framework (**Error! Reference source not found.**A). In contrast to prior reports (Caucheteux et al., 2023; Kumar et al., 2024), we found no evidence for systematic layer-wise progression in the spatial extent of significant predictions. Overall, most layer pairs exhibited no significant differences across the majority of ROIs, indicating broadly similar brain-model alignment patterns across layers. Also, there seems to be no cross-linguistic consistency in which pairs are significant. However, in Mandarin, structured deviations were observed for comparisons involving mid-level layers (approximately layers 5–6) and later layers (approximately layers 10–12), which consistently yielded a small subset of ROIs showing differences. What is worth noticing is that, across all three languages, neighboring layers showed the highest representational similarity, whereas similarity progressively decreased with increasing layer distance (Figure S4). This indicates that neural encoding scores stay stable despite the extracted mBERT representations changed systematically across model depth. Furthermore, to test whether the limited layer-wise differentiation was specific to the bidirectional masked-language architecture of mBERT, we repeated the layer-comparison analysis in English using two causal autoregressive models. GPT-2 showed similarly limited differentiation across layers, whereas Mistral exhibited more pronounced but spatially restricted layer-pair differences. Thus, the degree of layer differentiation varied across model architectures and scales, but neither autoregressive model showed a simple, systematic progression from earlier to higher-order cortical regions across model depth (Supplementary Figure S11). These findings suggest that brain alignment is largely stable across transformer layers.

Next, to address the issue of whether the layer performance reflects the processing hierarchy of language processing, we visualized the best-performing layer in each ROI (**Error! Reference source not found.**B). Across languages, early model layers yielded the strongest encoding in several regions, including the pSTG in Mandarin, frontal regions in English, and occipital regions in French. These patterns do not map cleanly onto canonical cortical hierarchies of language processing and show no consistency across languages. A similar pattern is observed in other percentiles (Figure S5-7). Within each language, the bootstrap layer generally agreed with the original group-level max score assignment (English: 92.1%; Mandarin: 88.5%; French: 91.3%). As confirmed in the bootstrap, preferred-layer robustness was not spatially consistent across languages. ROI-wise stability was essentially uncorrelated between language pairs (Pearson $r$ = .001–.022). Although isolated ROIs exhibited highly stable preferences within individual languages, these regions did not form a reproducible progression from earlier transformer layers in sensory systems to later layers in higher-order association cortex.

**Large-scale cortical organization is preserved rather than progressively reorganized across transformer depth**

Next, we investigated how average brain scores fluctuate across layers at a functional network level. To take into account layer-wise encoding performance, we first averaged the layer-wise brain scores in 7 functional networks and subcortical areas, as shown in Figure 5C. Across networks, brain scores exhibited a recognizable U shape over model depth. Notably, peak values were observed in the limbic network (max = 0.025), ventral attention network (max = 0.026), subcortical system (max = 0.035), and default mode network (max = 0.038), relative to consistently lower values in the visual network (max = 0.019), sensory-motor network (max =

0.019), and dorsal attention network (max = 0.016). This pattern is consistent with a cortical hierarchy of cognitive processing and with evidence that the ventral attention system preferentially supports semantic and conceptual aspects of language (Fujii et al., 2016; Saur et al., 2008). As seen in Figure 5D, on the other hand, this pattern is nearly universal across all 12 mBERT layers and three languages, showing no clear distinction for the predictive hierarchy seen with respect to model depth. This echoes the distributed layer-wise spatial patterns observed earlier at a whole-brain level in Figure 2.

### Contextualization shows unique contribution but not predictive advantage

To revisit prior reports (Caucheteux & King, 2022; Kumar et al., 2024; Mischler et al., 2024; Pasquiou et al., 2022) suggesting that contextual embeddings (e.g., BERT) capture activity in regions associated with higher-level linguistic and semantic processing compared to non-contextual embeddings (e.g., GloVe or limited-context embeddings), we used FastText as a contrasting static model. As Figure 6A indicated, both layer 12 performance using both HRF and FIR appear to be comparable with Fasttext prediction. One-sided sign-flip permutation tests for the hypothesis BERT > FastText were not significant (all $p > 0.95$), reflecting consistent effects in the opposite direction. As a complementary robustness analysis, we fit a pooled linear mixed-effects model to ROI-level scores, treating Model (mBERT vs FastText) as a fixed effect and including random intercepts for subject and ROI. Across all three languages, the fixed-effect estimate for Model (BERT − FastText) was negative and highly significant ($p < 10^{-40}$), confirming a higher average encoding performance for FastText when accounting for subject and ROI level variability consistent in direction with the permutation-based results reported. Interestingly, in another robustness-checking banded ridge regression model, when analyzing the unique variance, no parcel showed significant positive variance unique to L0 (Figure 6B). In contrast, 46 parcels showed significant positive variance unique to L12, indicating that although the contextualized representation provided predictive information beyond the initial token representation, the unique contribution doesn't necessarily capture the context-related areas.

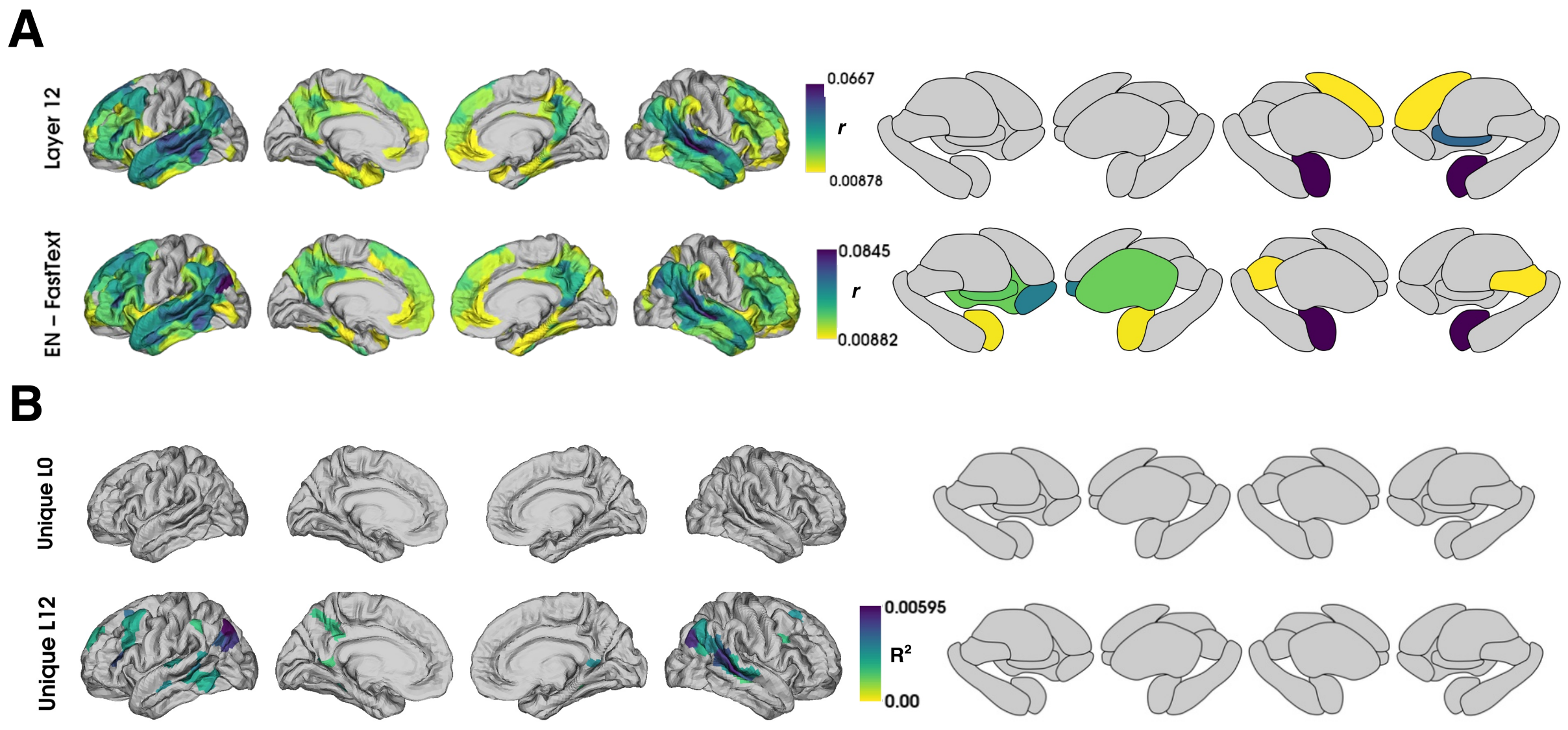

**Figure 6**: Comparison between L12 and Fasttext and unique contribution of contextualization. (A) Top: Significant predictions from the output layer of English mBERT. Bottom: Corresponding predictions from English FastText. The color bars are Pearson coefficients. (B) Unique predictive variance associated with mBERT layers 0 and 12. Colored parcels show the group-mean unique predictive ($R^2$) among parcels surviving ($q$<0.05). Gray parcels did not survive FDR correction.

**Surprisal and ID do not account for brain score profiles**

At the whole-brain level, brain scores showed a U-shaped profile across layers, similar to the ones seen across functional networks (Figure 5C), consistent with previous literature (Antonello et al., 2024; Kumar et al., 2024; Mischler et al., 2024; Oota et al., 2023; Zada et al., 2025). To further interpret these layer-wise patterns and the linguistic processes underlying LLM-based predictions of brain activity, we selected surprisal and ID as two indices of information prediction and compression, respectively, across model layers. Surprisal decreased progressively and showed a marked drop in the final layer, whereas ID peaked at the 10th layer, similar to existing reports (Cheng et al., 2025; Cheng & Antonello, 2024). Neither metric mirrored the brain score profile, failing to explain the superior predictive performance of the middle layers (Figure 7A-C). These findings suggest that the alignment between brain activity and LLM representations is not primarily driven by prediction accuracy or information compression.

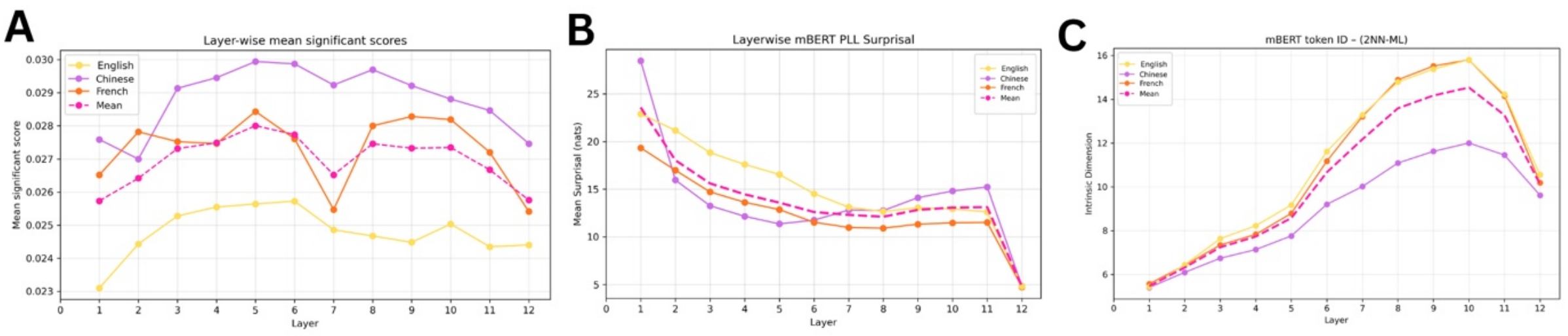

**Figure 7**: Layer-wise brain scores, surprisal, intrinsic dimensionality (ID), and their cross-lingual convergence. (C) Layer-wise mean brain scores were computed by aggregating and averaging scores across all significant ROIs, for Chinese (CN), English (EN), and French (FR). (D) Surprisals were likewise aggregated and averaged within each layer. (E) ID was estimated at each mBERT layer using the Two-NN method. Lower ID values indicate greater representational compression, while higher values reflect increased geometric complexity of the embedding space.

## DISCUSSION

The present study examined what aspects of brain–language model correspondence are empirically robust and what computational interpretations this robustness can support. Across Mandarin, English, and French, language-model representations predicted neural responses across a distributed set of cortical and subcortical regions, with substantial spatial overlap across languages. At the same time, neural predictivity showed limited systematic reorganization across transformer depth, and this pattern remained qualitatively similar in autoregressive models. Contextualized representations contributed unique predictive information beyond the non-contextual BERT input representation, but this contribution did not translate into a broad whole-brain advantage over static lexical-semantic embeddings or a reproducible higher-order cortical topography. Finally, neither surprisal nor intrinsic dimensionality reproduced the layer-wise profile of brain scores. Together, these findings distinguish the robustness of neural predictivity from stronger claims about computational equivalence between language models and the brain.

The spatial distribution of neural predictivity replicated several well-established findings from naturalistic language research. Across languages, significant encoding effects were observed in auditory and temporal regions, including STG and MTG, as well as IFG, IPL, PCC, precuneus, and mPFC. The involvement of trans-modal cortical regions is consistent with previous evidence that naturalistic language comprehension recruits distributed systems beyond the classical peri-sylvian language network, including the DMN and other regions associated with semantic integration, narrative construction, memory, and contextual updating (Caucheteux & King, 2022; Fernandino & Binder, 2024; Schrimpf et al., 2021; Simony et al., 2016; Zada et al., 2025; Zhou et al., 2025). We therefore interpret the distributed cortical effects primarily as a validation of the present encoding framework rather than as evidence for a previously unrecognized cortical language system. A principal extension of the present analysis is the systematic inclusion of subcortical structures. Several subcortical regions, including the caudate, hippocampus, thalamus, amygdala, and putamen, showed significant encoding performance across model layers and languages. These observations are consistent with prior evidence implicating the basal ganglia in semantic and sequencing-related processes and the hippocampus in contextual binding, event segmentation, referential processing, and the integration of information across extended temporal contexts (Blank et al., 2016; Dijksterhuis et al., 2024; Duff & Brown-Schmidt, 2012; Kurczek et al., 2013; Park et al., 2025; Tovar et al., 2024a, 2024b). The present results therefore show that language-model representations contain information predictive not only of cortical responses but also of activity in subcortical systems previously associated with memory, sequencing, and contextual integration.

The spatial overlap across Mandarin, English, and French further suggests that a substantial component of brain-LLM correspondence generalizes across typologically distinct languages. Shared effects were observed in cortical regions including pSTG, MTG, IFG, IPL, PCC, mPFC, and ACC, together with overlapping subcortical predictions. This pattern is consistent with recent multilingual studies reporting shared neural–model representational structure and cross-language transfer of encoding performance (Varda et al., 2025; Zada et al., 2025). At the same time, several language-specific differences were also present. These should not be attributed uniquely to linguistic typology. Differences in model training data, tokenization, translation choices, narration, and data acquisition may also contribute to cross-language variation. The cross-linguistic results are therefore most appropriately interpreted as evidence for robust spatial commonality in neural predictivity rather than as evidence that all three languages instantiate identical neural or computational representations.

A central question was whether transformer depth maps systematically onto cortical processing hierarchy. In contrast to reports of progressively deeper model layers preferentially predicting progressively higher-order cortical regions, we found limited evidence for such a one-to-one progression. Pairwise layer comparisons showed that most layer pairs differed in only a small proportion of ROIs, and preferred-layer maps did not reveal a reproducible transition from early layers in sensory cortex to later layers in transmodal association cortex. Several additional analyses argue against the possibility that this limited layer differentiation is simply an artifact of unstable layer selection or redundant hidden-state extraction. First, the CKA analysis showed that mBERT representations changed systematically across depth (Figure S7), therefore, similar neural encoding performance across layers cannot be explained by the extraction of effectively identical hidden states. Second, preferred-layer assignments were highly reproducible under participant bootstrap resampling within each language, yet the spatial pattern of layer stability showed essentially no correlation across languages. This indicates that the absence of a common cortical gradient was not simply due to an unstable group-level maximum, as the percentile-based sensitivity analyses led to a similar conclusion. Third, GPT-2 showed similarly limited differentiation across layers, whereas Mistral exhibited more pronounced but spatially restricted layer-pair differences. Thus, the degree of layer differentiation varied across model architectures and scales, but neither autoregressive model showed a simple, systematic progression from earlier to higher-order cortical regions across model depth (Figure S11-14). These autoregressive architectures differ substantially from mBERT in training objective, directionality, and depth. Hence, their similar encoding profiles make it less likely that the absence of a clear transformer-to-cortical hierarchy is explained solely by the bidirectional masked-language-model architecture of BERT.

These findings do not imply that transformer layers are functionally interchangeable, nor do they invalidate prior reports of layer-specific brain correspondence. Rather, they highlight that the notion of "hierarchy" has been operationalized in substantially different ways across studies. Some work compares the embedding layer with a selected deeper transformer layer (Caucheteux et al., 2022), others define hierarchy through the best-predicting layer at each region (Kumar et al., 2024), project model representations into a low-dimensional space (Antonello et al., 2021), or relate layer-dependent alignment to cortical distance from sensory regions (Mischler et al., 2024). These analyses address related but not identical questions. The present results therefore argue specifically against a simple, spatially reproducible one-to-one mapping between transformer depth and macroscale cortical hierarchy under the current naturalistic fMRI encoding framework. One possible explanation is that brain-relevant information is preserved across transformer depth even as the overall geometry of the representations changes. Transformer layers retain substantial lexical-semantic information rather than replacing earlier representations entirely (Jawahar et al., 2019; Kumar et al., 2024). If neural predictivity depends strongly on such preserved information, multiple layers may achieve similar encoding performance despite undergoing different internal transformations. This interpretation is also consistent with the present contextualization analyses, which indicate that contextual representations add information but do not produce a broad whole-brain superiority over static lexical-semantic representations. What is worth noticing is that at the network level, encoding performance was generally stronger in transmodal systems such as the DMN, limbic, and ventral attention networks than in primary visual and somatomotor systems. This distribution is broadly compatible with the large-scale organization of cortex described by macroscale functional gradients (Margulies et al., 2016). However, this network-level pattern should be distinguished from a transformer-layer hierarchy. The relative ranking of cortical networks was largely preserved across model layers, suggesting that model depth did not systematically shift neural predictivity from lower-order toward higher-order systems. Thus, the brain itself may exhibit hierarchical functional organization while the correspondence between transformer depth and that hierarchy remains weak or non-unique.

The revised banded ridge regression analyses also refine the interpretation of the contextual-versus-static comparison. In the original model-level comparison, FastText achieved equal or higher overall encoding performance than mBERT across the three languages. However, comparisons between independently trained models such as FastText and BERT confound contextualization with differences in architecture, training objective, vocabulary, tokenization, and dimensionality. The within-model comparison therefore provides a cleaner test of the contribution of contextualization. When the non-contextual BERT input embedding was compared with contextualized layer 12, the contextual representation achieved higher encoding performance. Moreover, the banded-ridge variance-partitioning analysis identified significant positive variance unique to layer 12 in 46 parcels, whereas no parcel showed significant positive variance unique to layer 0. These results demonstrate that contextualization does provide predictive information beyond the initial non-contextual token representation. At the same time, this additional information did not translate into a broad superiority over the static FastText baseline, nor did the unique contextual contribution form a reproducible topography selectively concentrated in canonical higher-order contextual regions. Such distinction is important because total predictive performance and unique predictive information are not equivalent quantities. A contextual representation can contain information not present in the model's input embedding while still failing to outperform a strong independently trained lexical-semantic representation at the whole-brain level. The present findings therefore do not support the stronger interpretation that contextualization is irrelevant to neural predictivity. Instead, they suggest that contextualization contributes additional information whose neural benefit is relatively selective rather than uniformly dominant. This result is compatible with prior evidence that static embeddings continue to explain substantial neural variance even when contextual models achieve higher aggregate performance (Goldstein et al., 2022).

Stimulus properties may also modulate the relative advantage of contextual representations. *The Little Prince* is a continuous naturalistic narrative but has relatively simple grammatical structure and a largely linear storyline. Under such conditions, lexical-semantic information may already explain a considerable proportion of the variance accessible to BOLD encoding models, whereas contextual advantages may become more pronounced in materials requiring stronger disambiguation, long-range dependencies, or conversational updating (Arora et al., 2020). This possibility remains speculative, however, because the present study includes only one narrative and cannot directly compare stimulus complexity across datasets.

The final set of analyses asked whether two model-internal quantities proposed as candidate explanations of brain-LLM alignment, surprisal and intrinsic dimensionality, could account for the observed layer-wise pattern of neural predictivity. Surprisal decreased progressively across model depth and showed a pronounced reduction in the final layer, whereas intrinsic dimensionality followed a different trajectory, peaking at later-intermediate layers. Neither profile reproduced the layer-wise brain-score curve. The dissociation between these trajectories places limits on surprisal and intrinsic dimensionality as explanations of the observed variation in neural predictivity. Importantly, this result should not be interpreted as evidence that predictive processing or representational compression are absent from either the brain or language models. Surprisal is one specific index of predictive uncertainty, and intrinsic dimensionality captures only one aspect of the organization of representational space. The present findings instead indicate that these particular layer-wise quantities are insufficient to explain why certain model representations predict neural activity better than others. This interpretation is consistent with recent work questioning whether next-word prediction or surprisal alone explains the neural performance of language models (Antonello & Huth, 2024; Cheng et al., 2026; Cheng & Antonello, 2024). Similarly, the relationship between intrinsic dimensionality and brain alignment appears to vary across architectures, model scales, and training stages (Cheng et al., 2025, 2026; Skean et al., 2025). The present dissociation therefore supports a more cautious view in which prediction and representational compression may contribute to model and brain representations without uniquely determining their correspondence.

After examining different accounts, the present results emphasize a distinction between predictive correspondence and computational equivalence. Encoding performance demonstrates that a model representation contains information that covaries systematically with neural activity. It does not establish that the model and the brain arrive at that information through the same sequence of computations. This distinction is particularly relevant given the combination of findings observed here. Neural predictivity was robust across languages, distributed across cortical and subcortical systems, and relatively stable across transformer depth and architectures. Contextualized representations contributed unique information beyond non-contextual BERT embeddings, yet this contribution did not yield a broad whole-brain superiority over static lexical-semantic representations. At the same time, neither surprisal nor intrinsic dimensionality reproduced the neural layer-wise profile. Holistically, these observations are more consistent with a substantial contribution from representational information that is preserved across model transformations than with a simple one-to-one computational homology between transformer depth and cortical processing hierarchy. This interpretation does not require that brains and language models implement identical computations. Instead, both systems may encode overlapping statistical and semantic structure present in natural language, allowing representations generated by very different architectures to remain predictive of neural responses.

Several limitations should constrain the interpretation of these findings. First, fMRI has limited temporal resolution because of the sluggish hemodynamic response. Rapid contextual updates and predictive computations may unfold at timescales that are poorly resolved by BOLD signals, potentially attenuating differences between model layers or architectures. Studies using ECoG and MEG have reported hierarchical effects that may be more sensitive to temporal processing dynamics (Goldstein et al., 2025; Mischler et al., 2024; Raugel et al., 2025). Our direct comparison between the canonical HRF and a lagged FIR model yielded qualitatively similar spatial prediction patterns, with slightly lower overall performance under the FIR specification (Figure S15), suggesting that the principal results are not strongly dependent on this particular hemodynamic modeling choice. Nevertheless, the present analysis does not exclude more subtle region-specific HRF differences. Second, cortical and subcortical regions differ in measurement quality. We therefore quantified tSNR separately to assess whether lower signal quality in subcortical structures could account for the observed encoding patterns. These analyses provide a data-quality control but do not eliminate all region-specific differences in vascular, anatomical, or susceptibility-related properties. Third, the cross-linguistic comparisons are affected by several sources of variability beyond linguistic typology. mBERT is trained on imbalanced multilingual corpora, and tokenization quality differs across languages (Acs et al., 2024; Pires et al., 2019). Moreover, the English, Mandarin, and French cohorts listened to different published translations of *The Little Prince*. Although the underlying narrative was shared, the three versions cannot be assumed to be matched at the exact lexical or syntactic level. Translation choices may alter word frequency, sentence structure, and semantic specificity, while narration and timing may introduce additional variability. Language-specific effects should therefore not be attributed uniquely to typological or neurocognitive differences. Fourth, the present analyses were performed on parcel-level fMRI time series. ROI-level modeling facilitates whole-brain comparisons and reduces dimensionality but inevitably averages over within-parcel spatial heterogeneity. However, voxels within the same ROI may differ in encoding strength or preferred model representation, and parcel averaging could attenuate localized layer-specific effects. Agreement across parcellation resolutions does not rule out meaningful voxel-level variation. Finally, the present conclusions concern a specific naturalistic narrative, a limited set of model architectures, and a particular family of encoding analyses. The absence of a simple transformer-to-cortical hierarchy here should therefore not be interpreted as evidence that hierarchical correspondences cannot emerge under other stimuli, model families, neural recording modalities, or operational definitions.

Overall, the present study shows that brain-LLM correspondence is spatially distributed and cross-linguistically robust, yet this robustness does not map straightforwardly onto transformer depth, contextualization, surprisal, or intrinsic dimensionality. Contextualized representations contribute unique predictive information beyond non-contextual token embeddings, but this contribution does not translate into a broad higher-order cortical advantage over static lexical-semantic representations. Across multiple model architectures, neural predictivity remains comparatively stable despite systematic changes in internal representations. These findings therefore caution against interpreting high brain scores as direct evidence that language models and the brain implement the same computational hierarchy. Instead, brain-LLM alignment may substantially reflect representational information that is preserved across model depth, architectures, and languages. Determining which components of this shared representational structure arise from genuinely shared computations, and which instead reflect common exposure to the statistical structure of natural language, remains an important goal for future work.

**Acknowledgements**

We thank all the members of the Grammar and Cognition Lab and all the colleagues for helpful discussions and feedback.

**Declaration of competing interest**

P.H. has received grants and honoraria from Novartis, Lundbeck, Mepha, Janssen, Boehringer Ingelheim, OM Pharma, and Neurolite outside of this work.

**Funding sources**

This research was supported by the European Research Council (ERC-2023-SyG, 101118756). N.Y. was supported by the FI Joan Oró fellowship (Generalitat de Catalunya). Views and opinions expressed are, however, those of the authors only and do not necessarily reflect those of the European Union or the Agency. Neither the European Union nor the granting authority can be held responsible for them.

## Data and code availability

Data is available at https://openneuro.org/datasets/ds003643/versions/2.0.5. All analysis code will be made available at https://github.com/NikkiYng/Cross-lingual-robustness-of-LLM-brain-alignment-and-its-computational-roots.

## Supplementary Methods

### Yeo-7 aggregation and within-atlas parcel-stability analysis

The spatial localization reported in the main analysis was defined by the Schaefer-400 cortical atlas. To assess whether its large-scale network-level pattern was driven by an individual parcel or a small number of extreme parcel estimates, we performed a representative within-atlas sensitivity analysis using the unthresholded English mBERT prediction scores across transformer layers 1-12. The 400 cortical parcels were assigned to their canonical Yeo-7 functional networks according to the Schaefer atlas labels. For each participant and transformer layer, we first averaged the unthresholded prediction accuracies across all parcels belonging to each Yeo network. Group-level network means were then calculated by averaging these participant-level network estimates. Between-participant uncertainty was quantified using 10,000 bootstrap samples in which participants were sampled with replacement. Percentile-based 95% intervals were defined by the 2.5th and 97.5th percentiles of the bootstrap distribution. We separately examined sensitivity to the sampling of parcels within each network. In each of 10,000 iterations, parcels were sampled with replacement within each Yeo network, after which the network means and ranking of the seven networks were recalculated. Because cortical parcels are spatially dependent, the resulting parcel-resampling intervals were interpreted as descriptive stability intervals rather than inferential confidence intervals. Finally, we performed a leave-one-parcel-out analysis. Each of the 400 cortical parcels was removed in turn, and we recomputed the whole-cortex mean prediction accuracy, the mean prediction accuracy of the parcel's assigned network, and the identity of the highest-scoring network. The maximum absolute change in the whole-cortex and within-network means and the proportion of parcel deletions that preserved the highest-scoring network are reported. These analyses test whether the large-scale results are sensitive to individual parcels within Schaefer-400.

### Temporal signal-to-noise ratio

We quantified temporal signal-to-noise ratio (tSNR) to assess whether differences in ROI-level data quality could confound comparisons between cortical and subcortical prediction accuracy.

tSNR was calculated from the denoised ROI time series before the normalization applied in the encoding-model pipeline. For each participant, run, and ROI, tSNR was defined as

$$tSNR_{s,k,r} = \frac{mean_{t(Y_{s,k,r,t})}}{SD_{t(Y_{s,k,r,t})}}$$

where ($s$) indexes participants, ($k$) indexes runs, ($r$) indexes ROIs, and ($t$) indexes time points. tSNR was calculated separately for each run to avoid introducing variance from between-run baseline differences. Run-level estimates were then averaged with equal weighting within each participant. For every participant, mean subcortical tSNR was calculated across the 16 Tian-S1 regions, whereas mean cortical tSNR was calculated across the 400 Schaefer parcels. Cortical and subcortical tSNR were compared within each language using paired-samples t tests across participants. Paired standardized effect sizes were calculated as Cohen's $d_z$, defined as the mean within-participant difference divided by the standard deviation of the within-participant differences. Paired Wilcoxon signed-rank tests were also performed as distribution-robust sensitivity analyses.

**Association between subcortical tSNR and prediction accuracy**

We next tested whether regional variation in subcortical encoding performance could be explained by regional variation in tSNR. Prediction accuracy was defined as the cross-validated Pearson correlation between predicted and observed fMRI responses in held-out runs. tSNR and prediction-accuracy matrices were matched by participant and ROI, and the analysis was restricted to the 16 Tian-S1 subcortical regions. The final mBERT layer was selected a priori as the primary encoding-model analysis. First, tSNR and prediction accuracy were averaged across participants within each subcortical ROI, producing 16 regional pairs per language. Their association was summarized using Spearman's rank correlation. Uncertainty in this regional-mean correlation was quantified using 10,000 participant-level bootstrap samples. In each bootstrap iteration, participants were sampled with replacement, regional means were recalculated, and the Spearman correlation across the 16 ROIs was recomputed. Second, we retained the participant-by-ROI structure by centering both tSNR and prediction accuracy within each participant and calculating a Spearman correlation across the resulting participant–ROI observations. Statistical significance was assessed using 10,000 within-participant permutation iterations. In each iteration, the tSNR ROI labels were randomly permuted separately within every participant, thereby preserving participant-level distributions while disrupting the correspondence between regional tSNR and prediction accuracy. Third, a Spearman correlation across the 16 ROIs was calculated separately for each participant. The mean and median of these participant-specific correlations were reported descriptively. To evaluate whether the result depended on the selected model layer, the regional-mean, subject-centered, and within-participant analyses were repeated for all 12 mBERT transformer layers. These analyses were interpreted as signal-quality sensitivity tests. A nonsignificant tSNR association was not treated as proof that prediction accuracy was independent of all aspects of data quality.

### Comparison of HRF- and FIR-based temporal models

To assess whether the temporal alignment model influenced encoding performance, we compared English mBERT layer-12 prediction accuracies obtained using the canonical hemodynamic response function (HRF) with those obtained using a finite impulse response (FIR) model. The FIR model included five delays spanning 0–8 s in 2s increments. Both analyses used the same participants, cross-validation procedure, and 416-region parcellation. Analyses were performed on the unthresholded participant-by-region prediction-accuracy matrices, and all paired differences were defined as HRF minus FIR. At each region, the mean paired difference was tested against zero using a two-sided sign-flip permutation test with 10,000 permutations. Permutation $p$ values were corrected across the 416 regions using the Benjamini–Hochberg false-discovery-rate procedure ($q$ < 0.05), and significant regions were classified according to the direction of the mean difference. For the global comparison, prediction accuracy was first averaged across the 416 regions within each participant. The participant-level global differences were evaluated using a one-sided sign-flip permutation test testing HRF > FIR and a one-sided paired-samples t test; paired effect size was quantified using Cohen's $d_z$,. As a complementary pooled analysis, we fitted a linear mixed-effects model to all participant-by-region observations, with temporal model (FIR as the reference level and HRF as the comparison level) as a fixed effect and random intercepts for participant and region.

### Compute layer-wise difference for generative models

To assess whether regional prediction accuracy differed across the layers of the autoregressive models, we conducted paired permutation tests separately for GPT-2 and Mistral. For every pair of layers within each model and each ROI, we calculated the within-participant difference in prediction accuracy and used the mean difference across participants as the test statistic. Statistical significance was assessed using a two-sided paired sign-flip permutation test with 10,000 permutations. On each permutation, the sign of each participant's difference was independently reversed with equal probability. This procedure is equivalent to randomly exchanging the two layer labels within participants and therefore preserves the dependence among repeated measurements while testing the null hypothesis that the two layers are exchangeable. Monte Carlo $p$ values were calculated independently for each ROI. For each layer pair, the resulting 416 $p$ values were corrected across ROIs using the Benjamini-Hochberg false-discovery-rate procedure at $q$ = 0.05. This correction was performed separately for each layer pair. For each model and layer pair, we report the number and proportion of ROIs exhibiting a significant difference after FDR correction.

### Comparison of FastText, mBERT L0 and L12

To determine whether encoding performance differed between static FastText embeddings and mBERT's input embeddings (L0), and between mBERT's input and final contextualized representations (L0 and L12, respectively), we performed subject-wise paired comparisons of ROI-level encoding scores in the English dataset. For each comparison and ROI, we calculated the within-subject difference in Pearson correlation-based encoding scores (FastText minus L0 or L0 minus L12), with the mean difference across subjects serving as the observed test statistic. Statistical significance was assessed using paired sign-flip permutation tests. On each of 100,000 Monte Carlo permutations, the sign of every subject's difference was independently and randomly inverted, which is equivalent to exchanging the two model labels within subjects under the null hypothesis of exchangeability. Two-sided permutation $p$ values were calculated as the proportion of permuted absolute mean differences at least as large as the observed absolute mean difference, using a plus-one correction for both the numerator and denominator. Tests were performed independently for all 416 ROIs, and the resulting $p$ values were corrected across ROIs using the Benjamini–Hochberg false discovery rate procedure ($q < 0.05$). Additionally, as a complementary analysis of the overall difference across ROIs, we fitted a separate pooled linear mixed-effects model for each model comparison. ROI-level encoding scores were entered as the dependent variable and model identity as a fixed effect. Random intercepts were included for subject and ROI to account for repeated observations within subjects and variability across brain regions.

## Supplementary Results

### Stability of the English Yeo-7 networks

The unthresholded English mBERT layer-12 results showed a structured network-level profile (Table S1). Prediction accuracy was highest in the default-mode network, followed by the control network and visual network. Lower estimates were observed in the salience/ventral-attention network, somatomotor network, limbic network, and dorsal-attention network. The network ordering was stable to parcel resampling. At layer 12, the default-mode network was the highest-scoring network in all 10,000 within-network parcel-bootstrap iterations. The same result was observed at every examined transformer layer. In the leave-one-parcel-out analysis, removing any individual cortical parcel changed the layer-12 whole-cortex mean by no more than 0.00013, corresponding to 0.97% of the observed whole-cortex mean of $r = 0.01371$. The maximum absolute change in any within-network mean was 0.00085. The default-mode network remained the highest-scoring network after every one of the 400 possible single-parcel deletions. Across layers 1–12, the maximum absolute change in the whole-cortex mean ranged from 0.00013 to 0.00016, and the identity of the highest-scoring network was preserved after every parcel deletion. These results indicate that the large-scale English network-level pattern was not attributable to one Schaefer-400 parcel or a small number of extreme parcel estimates.

### Mean subcortical tSNR exceeded mean cortical tSNR

Subcortical tSNR was not lower than cortical tSNR in any of the three language datasets. Instead, mean subcortical tSNR exceeded mean cortical tSNR in every included participant. Paired comparisons showed large cortical-subcortical differences in English, Chinese, and French, with the same conclusion obtained from the corresponding nonparametric sensitivity tests (Table S2). These results indicate that the subcortical findings cannot be attributed to a general reduction in ROI-level temporal signal quality relative to the cortex.

**No observed relationship between subcortical tSNR and prediction accuracy**

No language showed evidence that subcortical regions with higher prediction accuracy simply had higher regional tSNR. For the final mBERT layer, the association between the 16 regional mean tSNR values and regional prediction accuracy was negative in English and weak in Chinese and French. When the participant-by-ROI structure was retained, the subject-centered associations were small and nonsignificant in all three datasets (Table S3). The layerwise sensitivity analysis yielded the same overall pattern. Across all mBERT layers and languages, subject-centered correlations ranged from -0.159 to 0.031, with no consistent positive association between regional tSNR and prediction accuracy. Together, these analyses argue against the specific confounding account that the subcortical encoding results were driven by regions with better temporal signal quality.

**HRF-based models outperformed FIR-based models**

English mBERT layer-12 prediction accuracy was higher with the canonical HRF model than with the FIR model. Mean prediction accuracy across participants and regions was 0.01412 for HRF and 0.00756 for FIR, corresponding to a mean paired difference of 0.00656. The participant-level global comparison was significant in both the sign-flip permutation test (one-sided $p$ = 0.0021) and the paired-samples t test, $t$ = 2.93, one-sided $p$ = 0.0026, Cohen's $d_{z}$,= 0.43. At the regional level, 46 of 416 regions survived correction in the two-sided permutation analysis (Benjamini–Hochberg FDR, $q < 0.05$). All 46 were cortical parcels showing higher prediction accuracy for HRF than for FIR; no subcortical region showed a significant difference, and no region showed significantly higher prediction accuracy for FIR. The pooled mixed-effects analysis yielded the same direction of effect (HRF minus FIR: $\beta$ = 0.00656, SE = 0.00032, $z$ = 20.24, two-sided $p = 4.53 \times 10^{-}91$). Together, these results indicate that, for English mBERT layer 12, the canonical HRF alignment produced higher prediction accuracy than the tested five-delay FIR specification, with the FDR-significant regional differences in the cortex.

**Limited layer-wise differentiation in generative models**

Pairwise analyses revealed limited layer-wise differentiation in GPT-2 and more pronounced, although spatially restricted, differentiation in Mistral (Figure S21). For GPT-2, 92.4% layer pairs showed no significant regional differences after FDR correction. Each of the remaining five comparisons identified a difference in only one of the 416 ROIs, corresponding to 0.24% of the analyzed regions. The median proportion of significantly different ROIs across GPT-2 layer pairs was therefore zero, and no comparison involved more than one significant ROI. For Mistral, 75.4% showed no significant regional differences, whereas 24.6% contained at least one significantly different ROI. The median proportion of significantly different ROIs across all Mistral layer pairs was nevertheless zero, indicating that most comparisons showed no regional differences. The largest difference was observed between Mistral layers 1 and 14, for which 58 of the 416 ROIs (13.9%) differed significantly. Thus, regional prediction accuracy was highly similar across GPT-2 layers, whereas Mistral exhibited greater depth-dependent differentiation concentrated in a subset of layer comparisons and brain regions.

### FastText outperformed mBERT L0 in encoding performance

Across subjects and ROIs, the mean encoding score was 0.01968 for FastText and 0.01572 for mBERT L0, corresponding to an average FastText-minus-L0 difference of 0.00396. Mean ROI-level differences favored FastText in 283 of the 416 ROIs. The two-sided paired sign-flip analysis identified 24 ROIs with a significant difference after FDR correction, all of which showed higher encoding scores for FastText than for L0 (minimum $q = 0.00416$). The pooled linear mixed-effects model yielded a consistent result. The fixed-effect estimate for FastText minus L0 was positive and significant ($\beta = 0.00396$, SE = 0.000367, $z = 10.77$, $p = 4.65 \times 10^{-27}$, 95% CI [0.00324, 0.00468]). Thus, both the ROI-wise paired analysis and the pooled robustness analysis indicated higher average encoding performance for FastText than for mBERT's static input embeddings.

### mBERT L0 and L12 show comparable encoding performance

Across subjects and ROIs, the mean encoding score was 0.01572 for L0 and 0.01412 for L12, corresponding to an average L0-minus-L12 difference of 0.00160. Mean ROI-level differences favored L0 in 249 of the 416 ROIs. No individual ROI survived correction in the two-sided paired sign-flip analysis (minimum $q = 0.0541$), providing no FDR-corrected evidence for a localized difference between L0 and L12. The pooled linear mixed-effects model nevertheless detected a small positive overall effect for L0 relative to L12 ($\beta = 0.00160$, SE = 0.000356, $z = 4.49$, $p = 6.99 \times 10^{-6}$, 95% CI [0.000903, 0.00230]). This pooled effect indicates a modest average advantage for L0 across the full set of ROIs, but it should not be interpreted as evidence that any particular ROI differed after correction for multiple comparisons.

### Supplementary Tables

**Table S1:** English mBERT layer-12 prediction accuracy within the seven Yeo cortical networks

| Network | Mean *r* | Participant-bootstrap 95% interval |
|---|---|---|
| Default mode | 0.0264 | [0.0216, 0.0313] |
| Control | 0.0150 | [0.0102, 0.0198] |
| Visual | 0.0129 | [0.0080, 0.0175] |
| Salience/ventral attention | 0.0104 | [0.0050, 0.0158] |
| Somatomotor | 0.0087 | [0.0043, 0.0132] |
| Limbic | 0.0058 | [0.0000, 0.0116] |
| Dorsal attention | 0.0043 | [-0.0014, 0.0099] |

*Note.* Network means were calculated from the complete unthresholded participant-by-parcel prediction-accuracy matrix. Values are group means of participant-level network averages. The 95% intervals quantify between-participant uncertainty.

**Table S2:** Cortical and subcortical temporal signal-to-noise ratio across language datasets

| Language | Subcortical tSNR, mean (*SD*) | Cortical tSNR, mean (*SD*) | Paired difference | 95% CI | t(df) | Cohen's $d_z$, |
|---|---|---|---|---|---|---|
| English | 372.01 (61.08) | 294.62 (45.76) | 77.39 | [68.97, 85.80] | 18.51 (46) | 2.70 |
| Chinese | 460.57 (49.19) | 342.77 (40.55) | 117.79 | [106.59, 128.99] | 21.40 (33) | 3.67 |
| French | 491.22 (58.03) | 373.74 (58.62) | 117.48 | [105.46, 129.50] | 20.05 (27) | 3.79 |

*Note.* tSNR was calculated separately for each ROI and run as the temporal mean divided by the temporal standard deviation; run-level values were then averaged within participant. Subcortical summaries were calculated across 16 Tian-S1 regions and cortical summaries across 400 Schaefer parcels. Positive paired differences indicate higher subcortical tSNR. Cohen's *dz* is the standardized within-participant difference.

**Table S3:** Association between subcortical tSNR and mBERT Layer 12 prediction accuracy

| Language | ROI-mean Spearman *ρ* | Bootstrap 95% CI | Subject-centered Spearman *ρ* | Permutation *p* | Mean within-participant *ρ* | Median within-participant *ρ* |
|---|---|---|---|---|---|---|
| English | -0.709 | [-0.824, -0.168] | -0.044 | .242 | -0.035 | 0.009 |
| Chinese | -0.138 | [-0.541, 0.238] | -0.021 | .635 | -0.005 | -0.046 |
| French | -0.065 | [-0.441, 0.312] | -0.047 | .329 | -0.052 | -0.022 |

*Note.* Prediction accuracy was defined as the cross-validated Pearson correlation between predicted and observed responses in held-out runs. ROI-mean correlations were calculated across 16 Tian-S1 regions after averaging both measures across participants; 95% intervals were obtained by resampling participants 10,000 times. Subject-centered correlations retained participant-by-ROI observations after centering both measures within participant. Permutation p values were obtained from 10,000 within-participant ROI-label permutations. The final two columns summarize correlations calculated separately across the 16 ROIs for each participant.

## Supplementary Figures

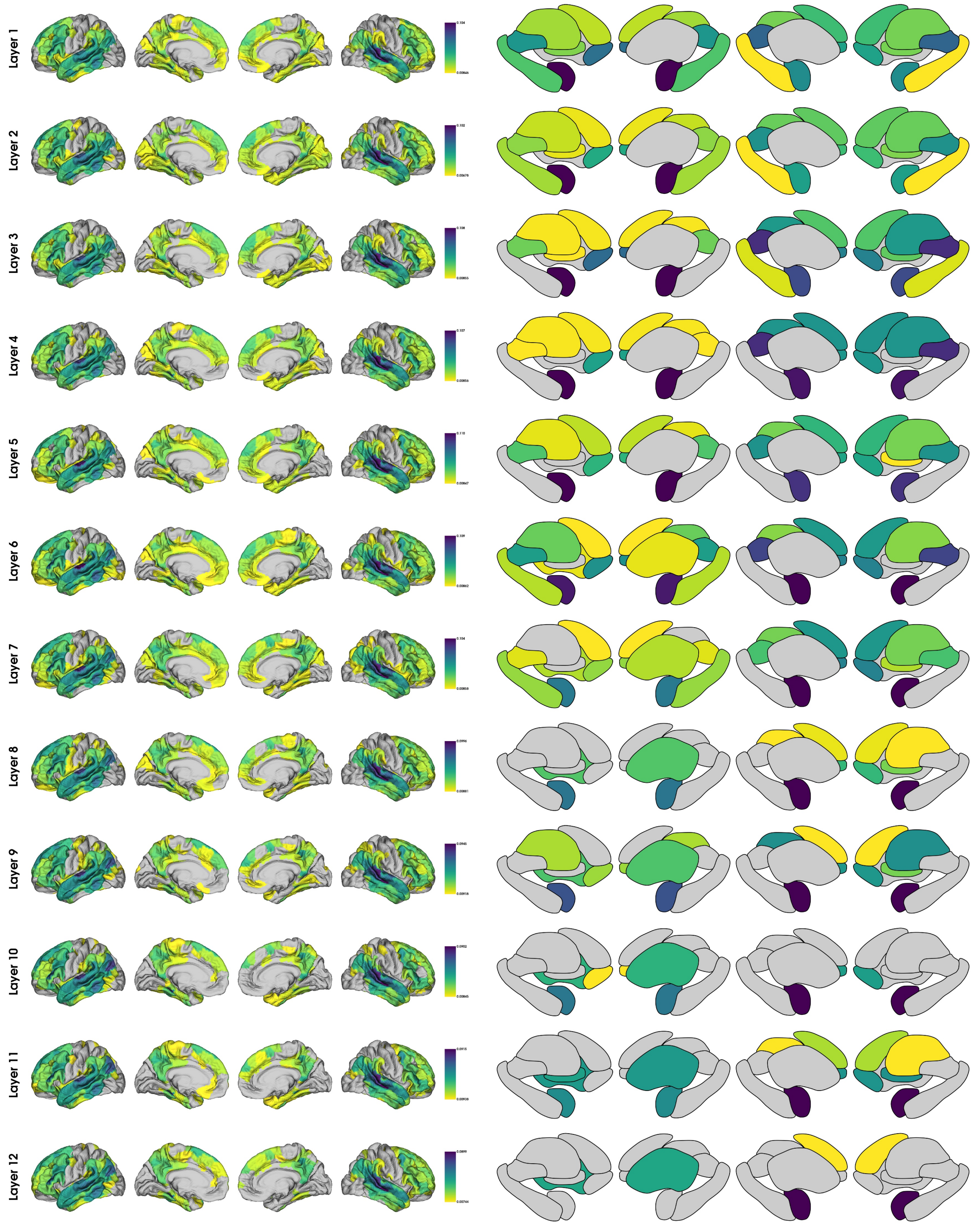

**Figure S8**: Cortical and subcortical prediction effects across layers for English mBERT. Only regions with significant brain scores are colored. Color bar indicates the highest and lowest significant score of this layer. Statistical significance was assessed using one-sample t-tests of subject-level prediction scores against zero (FDR-corrected, $p < 0.05$).

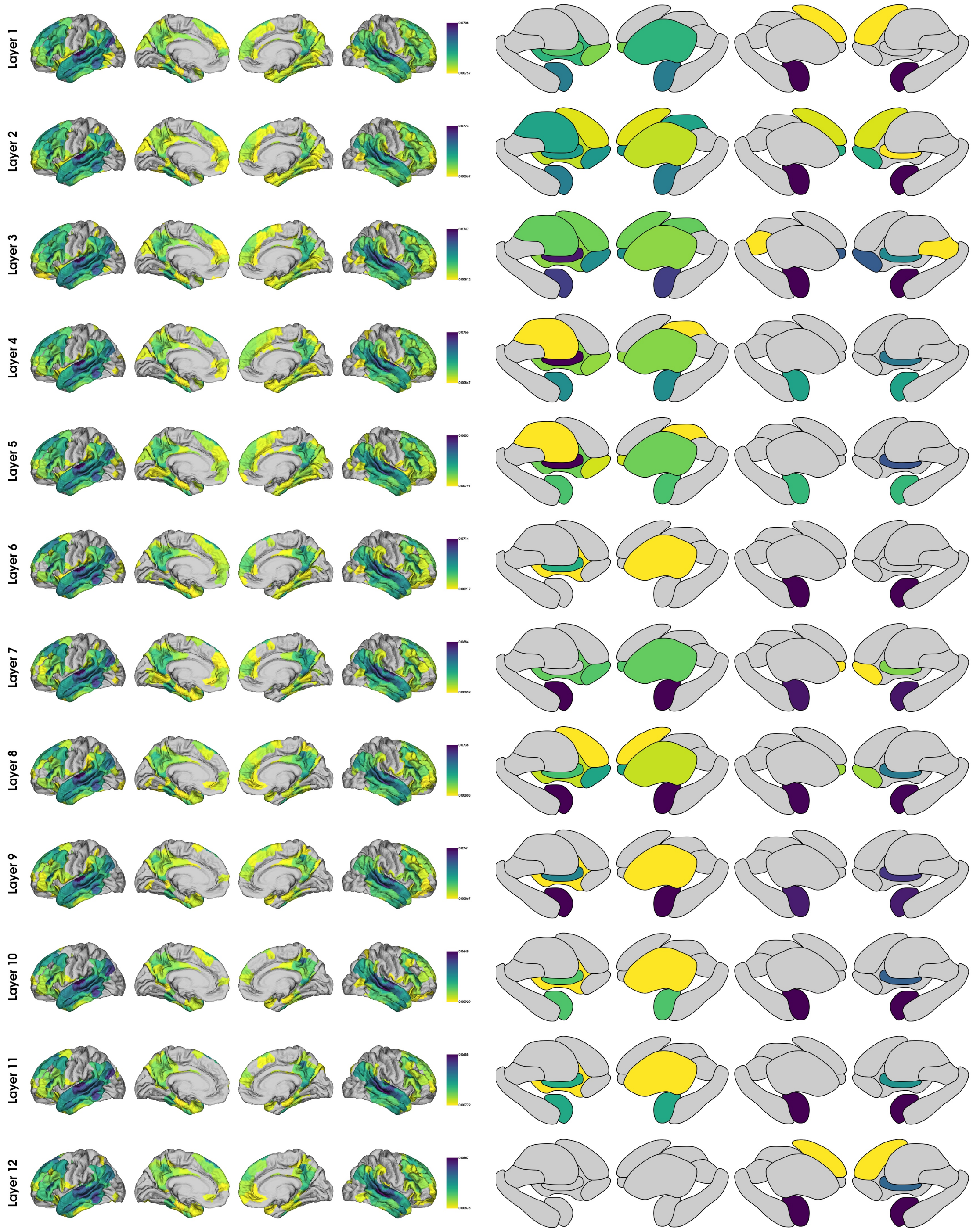

**Figure S9**: Cortical and subcortical prediction effects across layers for Mandarin mBERT. Only regions with significant brain scores are colored. Color bar indicates the highest and lowest significant score of this layer. Statistical significance was assessed using one-sample t-tests of subject-level prediction scores against zero (FDR-corrected, $p < 0.05$).

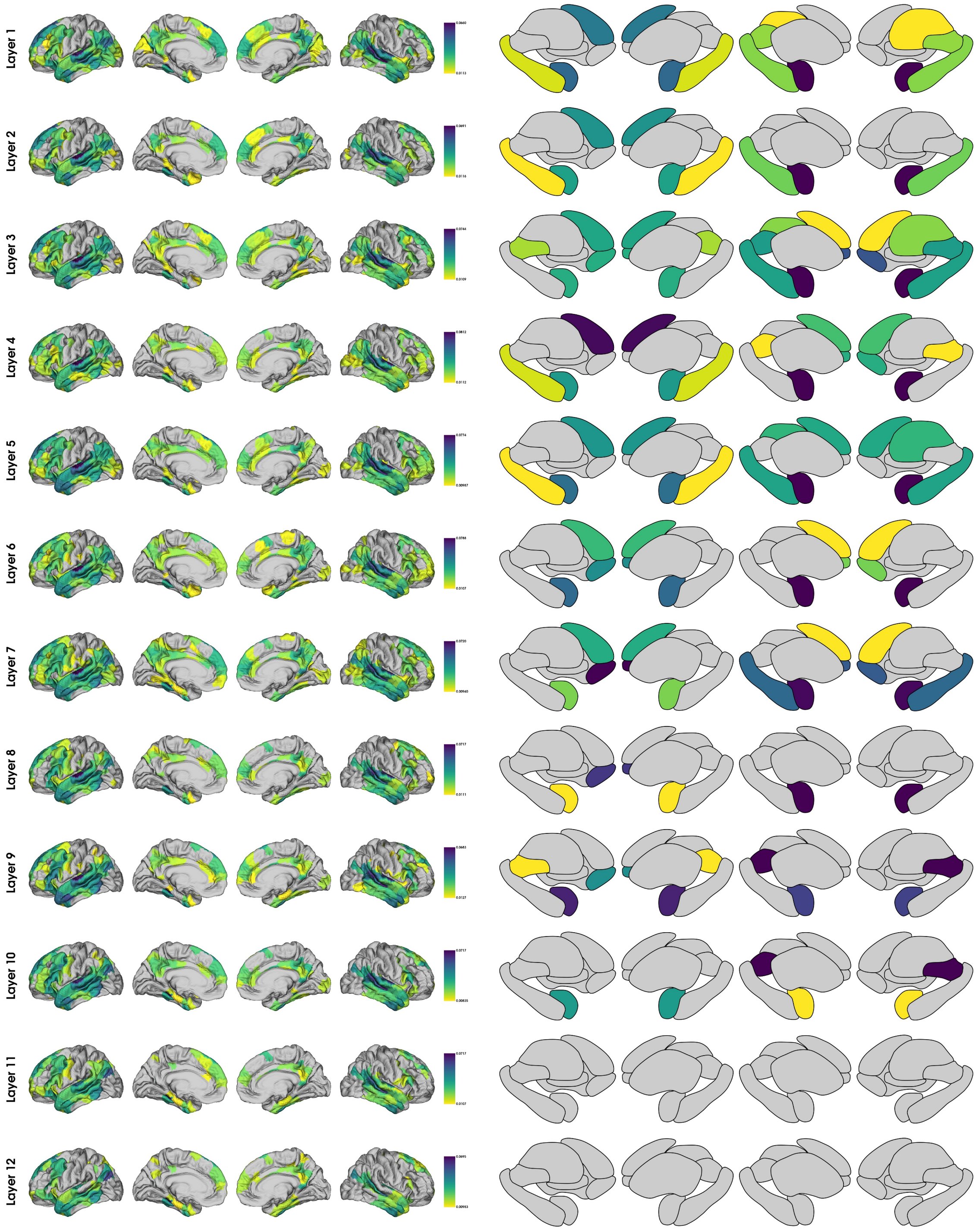

**Figure S10**: Cortical and subcortical prediction effects across layers for French mBERT. Only regions with significant brain scores are colored. Color bar indicates the highest and lowest significant score of this layer. Statistical significance was assessed using one-sample t-tests of subject-level prediction scores against zero (FDR-corrected, $p < 0.05$).

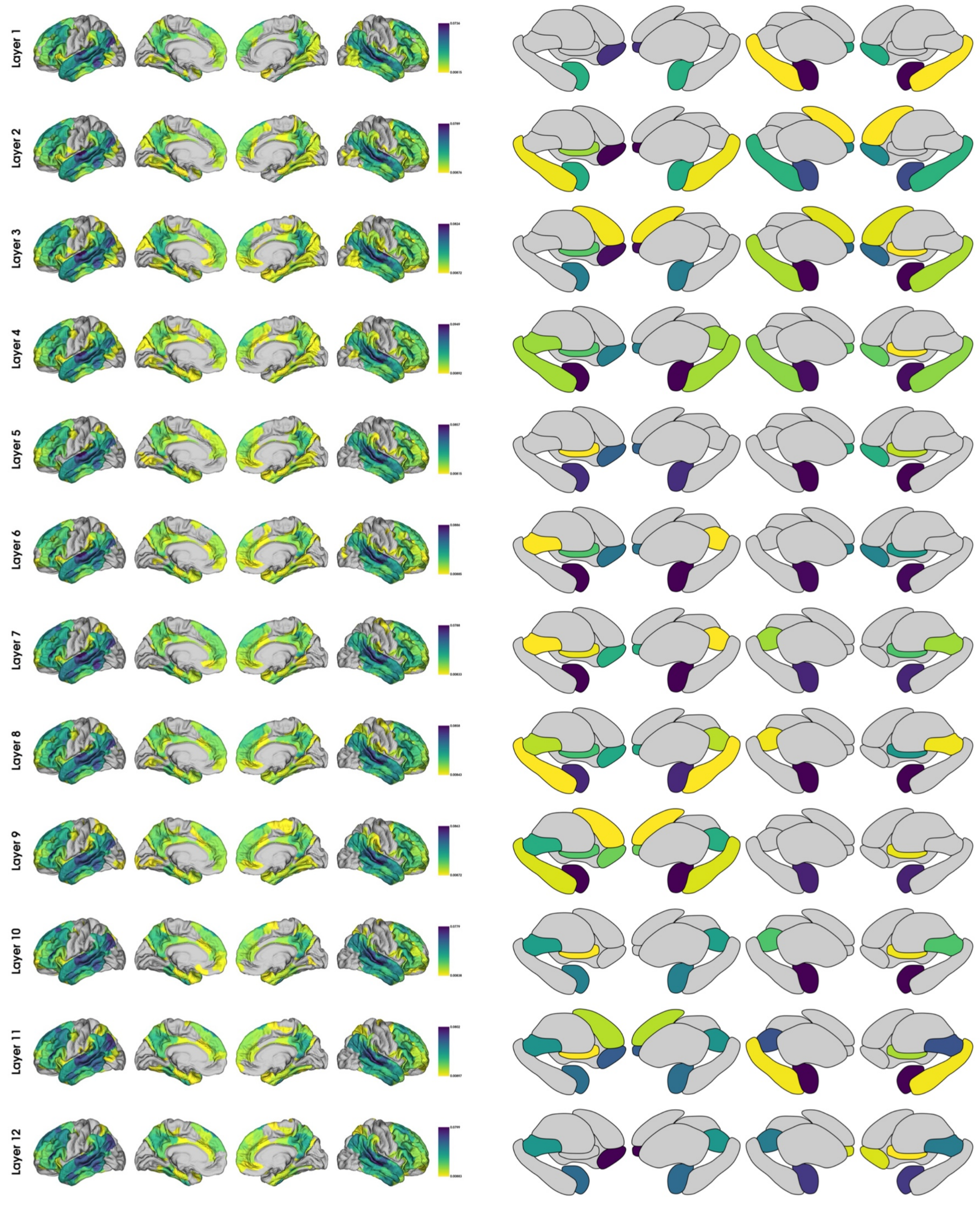

**Figure S11:** Cortical and subcortical prediction effects across layers for English BERT. Only regions with significant brain scores are colored. Statistical significance was assessed using one-sample t-tests of subject-level prediction scores against zero (FDR-corrected, $p < 0.05$).

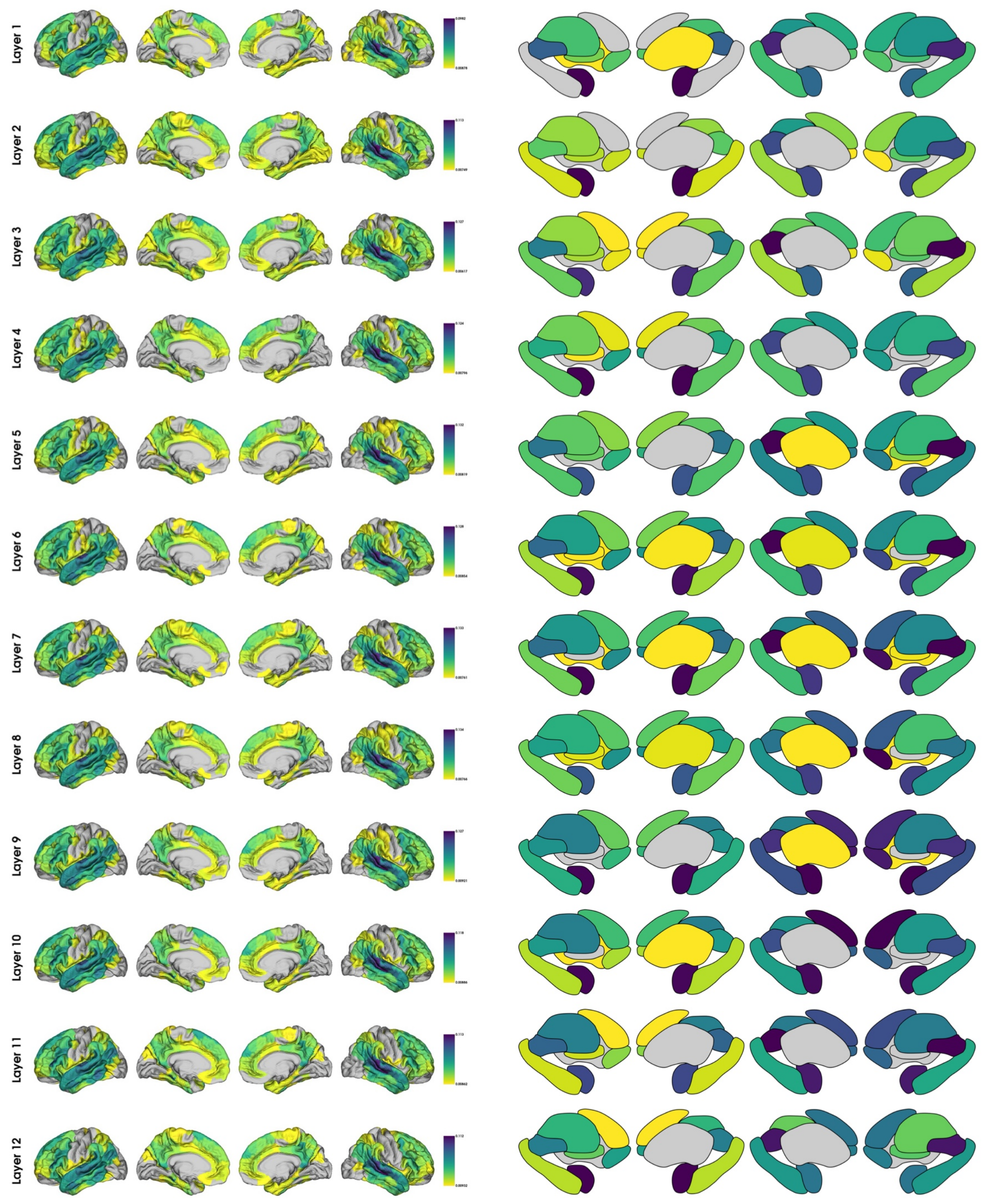

**Figure S12:** Cortical and subcortical prediction effects across layers for Mandarin BERT. Only regions with significant brain scores are colored. Statistical significance was assessed using one-sample t-tests of subject-level prediction scores against zero (FDR-corrected, $p < 0.05$).

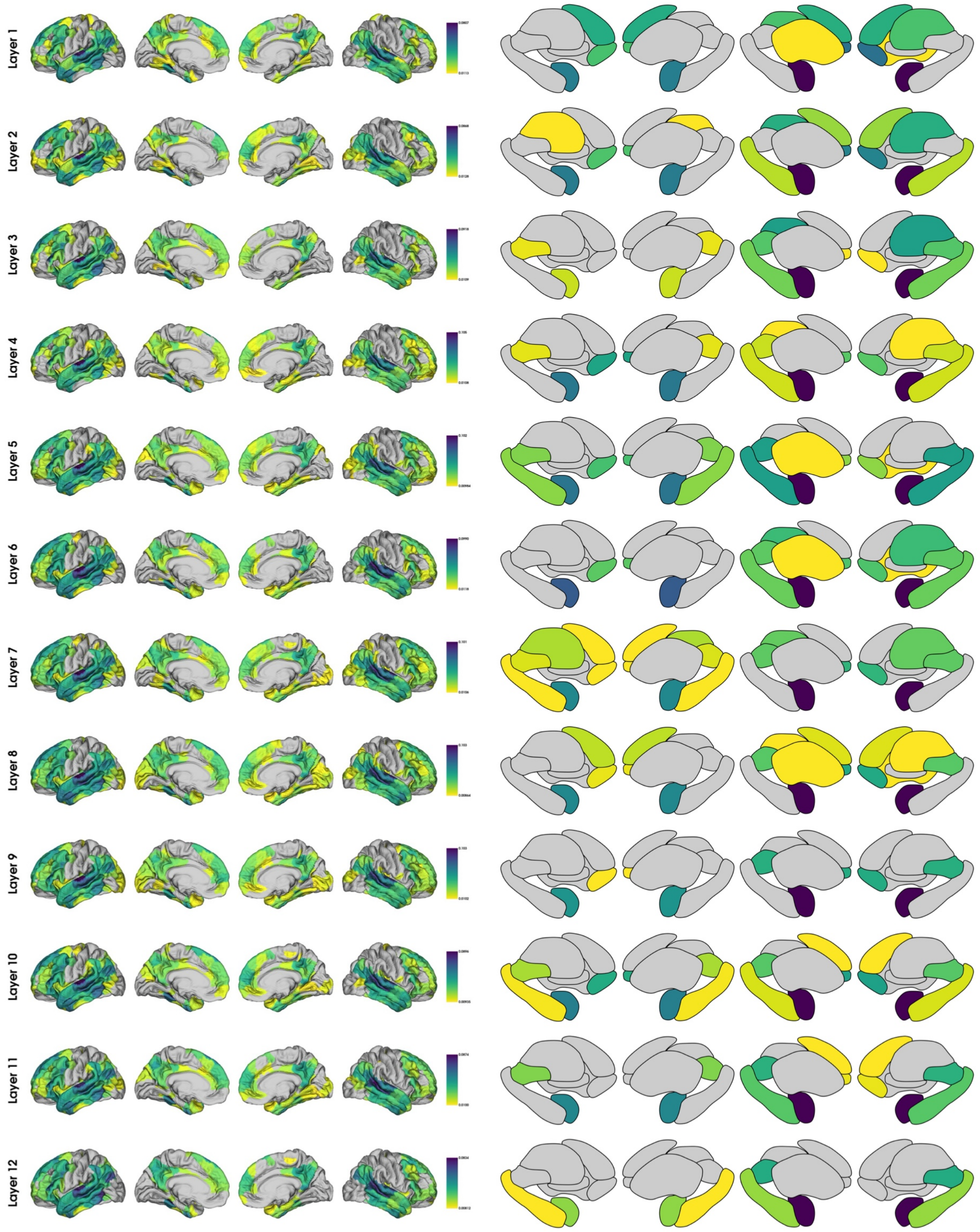

**Figure S13:** Cortical and subcortical prediction effects across layers for French BERT. Only regions with significant brain scores are colored. Statistical significance was assessed using one-sample t-tests of subject-level prediction scores against zero (FDR-corrected, $p < 0.05$).

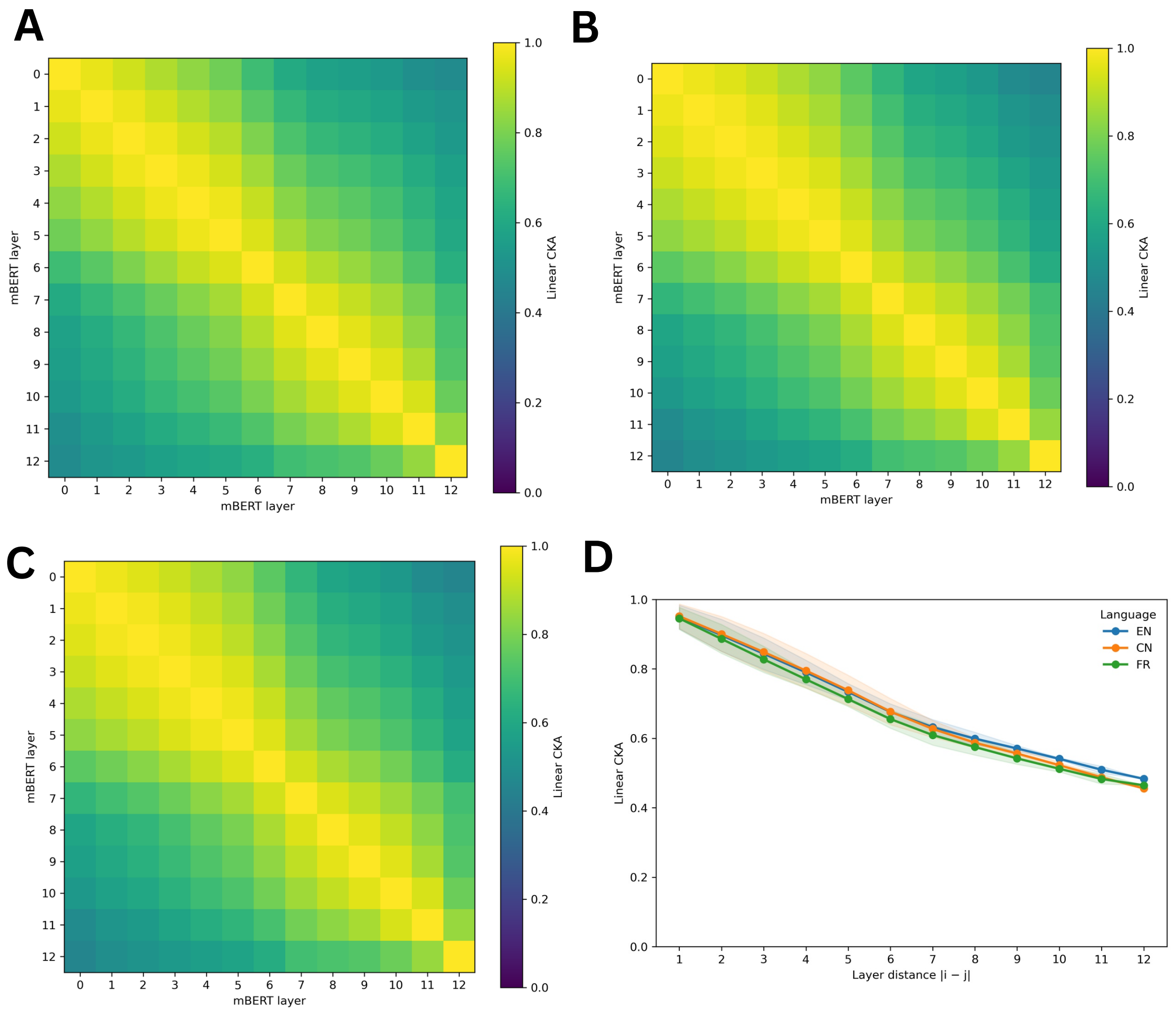

**Figure S14:** Layer-wise representational similarity of mBERT embeddings across languages. (A-C) Pairwise linear centered kernel alignment (CKA) between word-level representations from the mBERT embedding layer (layer 0) and the 12 Transformer blocks (layers 1-12) for (A) English, (B) Mandarin Chinese, and (C) French. Each cell represents the linear CKA between two layers computed across all aligned word occurrences within that language. Higher CKA values indicate greater similarity in representational structure. (D) Mean linear CKA as a function of absolute layer distance, *(|i-j|),* for the three languages. At each distance, CKA values were averaged across all layer pairs separated by that number of layers. Shaded regions indicate ±1 standard deviation across the corresponding layer pairs. All three languages showed a closely overlapping decline in representational similarity with increasing layer distance, indicating systematic transformation of mBERT representations across model depth.

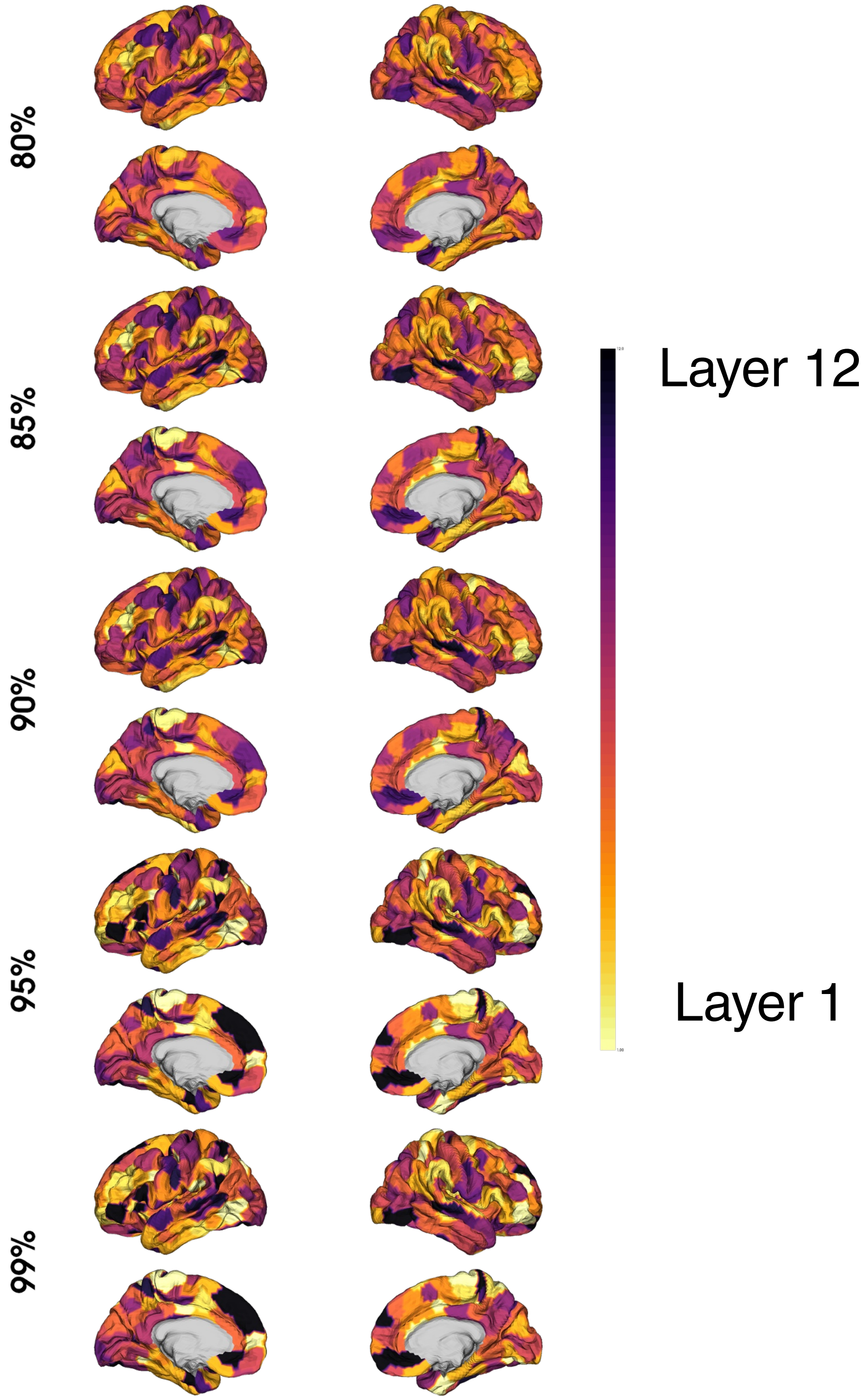

**Figure S15:** Percentile sensitivity of preferred-layer estimates for English mBERT. Cortical maps show the mean index of the top-performing mBERT layers for each parcel at the 80th, 85th, 90th, 95th, and 99th percentile thresholds, displayed from top to bottom. Percentile thresholds were calculated separately within each cortical parcel across Layers 1–12.

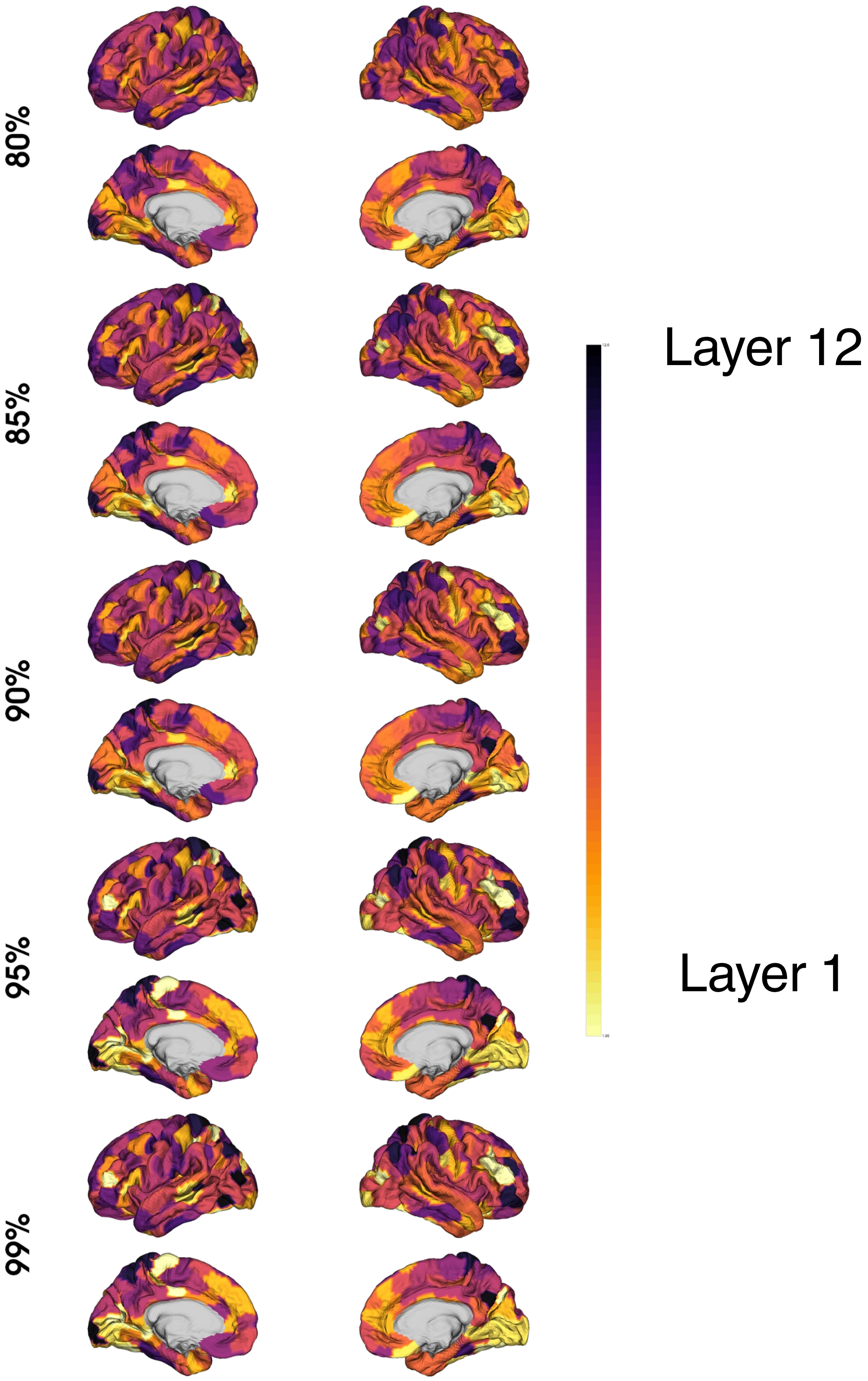

**Figure S16:** Percentile sensitivity of preferred-layer estimates for Madarin mBERT. Cortical maps show the mean index of the top-performing mBERT layers for each parcel at the 80th,

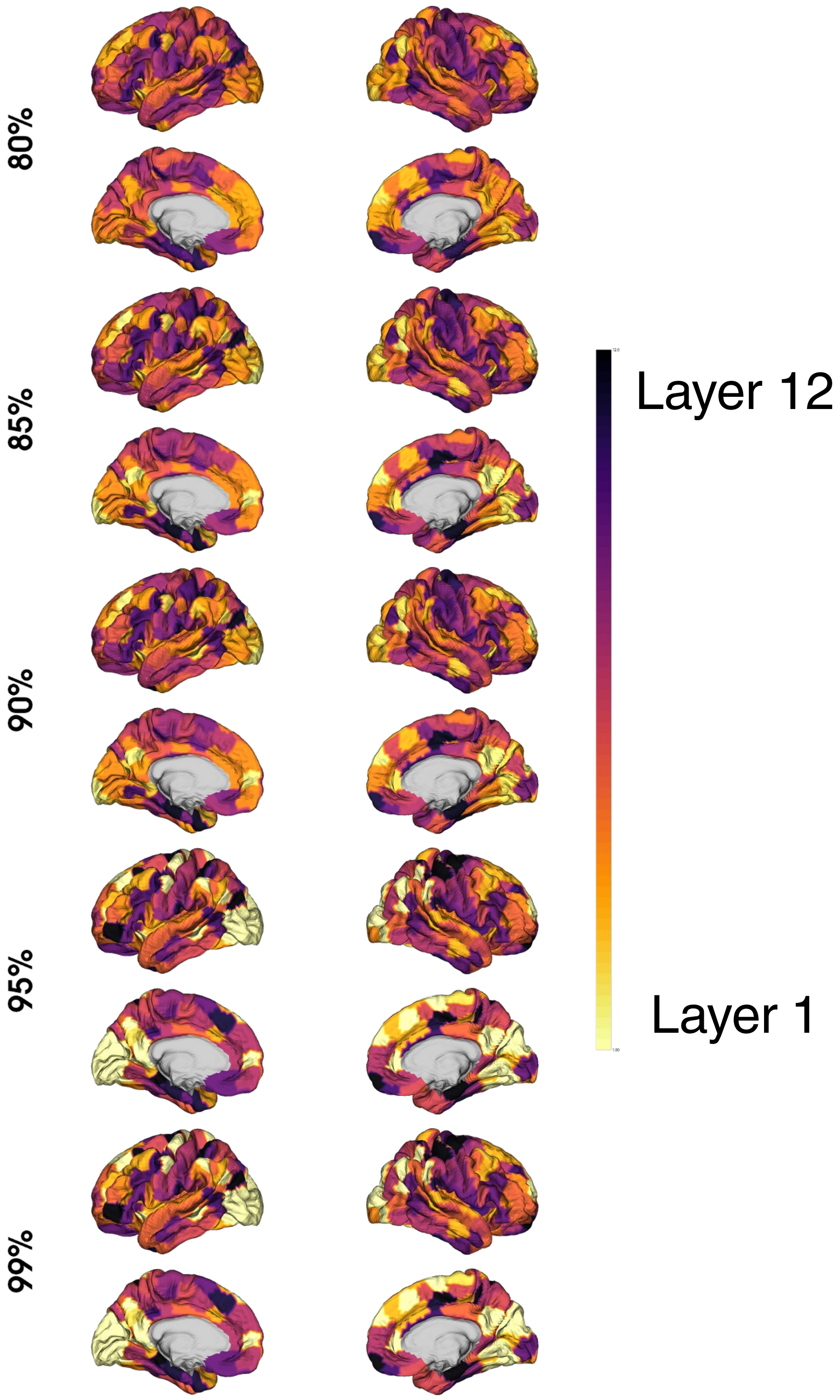

**Figure S17:** Percentile sensitivity of preferred-layer estimates for French mBERT. Cortical maps show the mean index of the top-performing mBERT layers for each parcel at the 80th, 85th, 90th, 95th, and 99th percentile thresholds, displayed from top to bottom. Percentile thresholds were calculated separately within each cortical parcel across Layers 1–12.

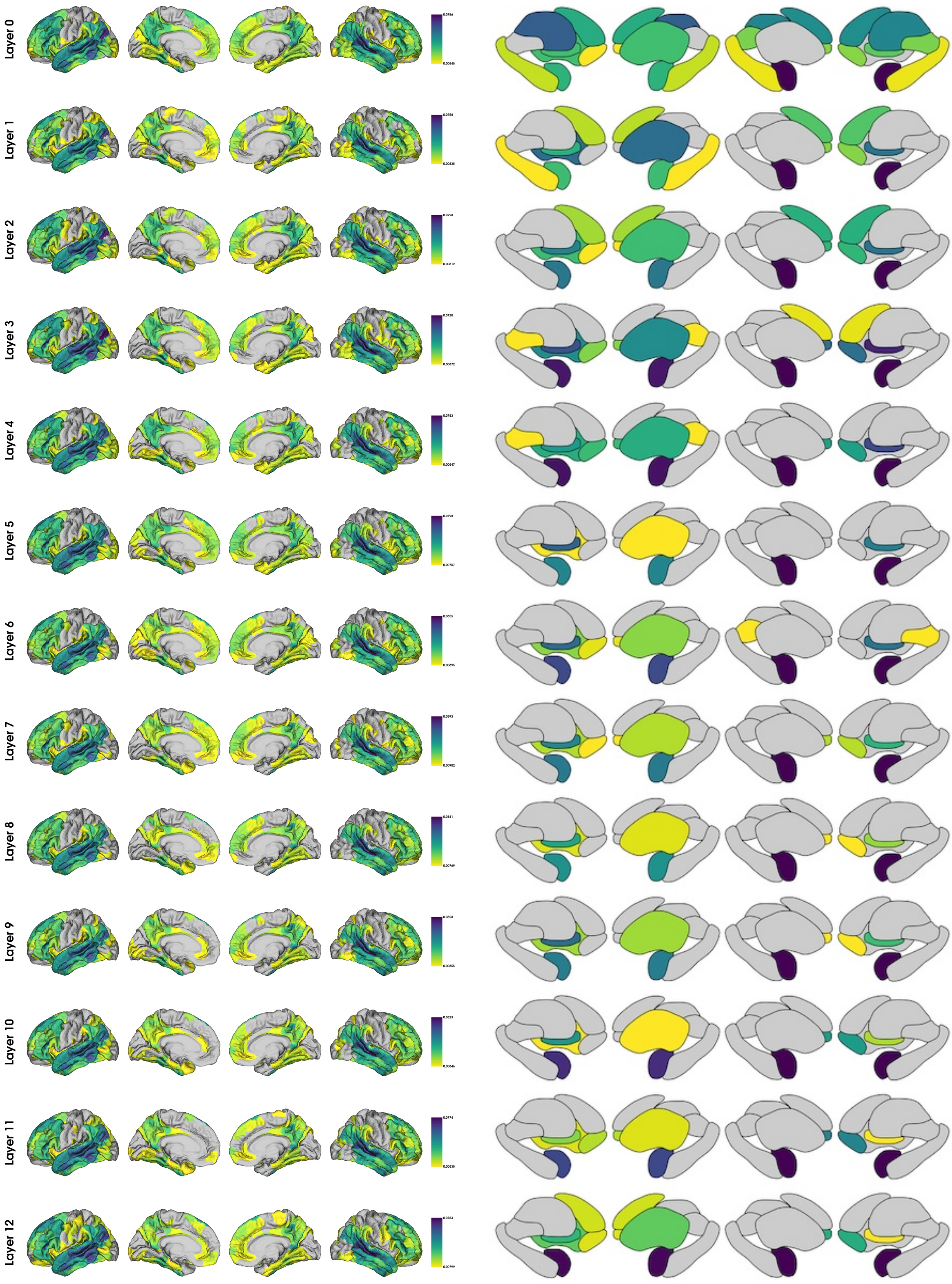

**Figure S18:** Cortical and subcortical prediction effects across layers for GPT2 English. Only regions with significant brain scores are colored. Statistical significance was assessed using one-sample t-tests of subject-level prediction scores against zero (FDR-corrected, $p < 0.05$).

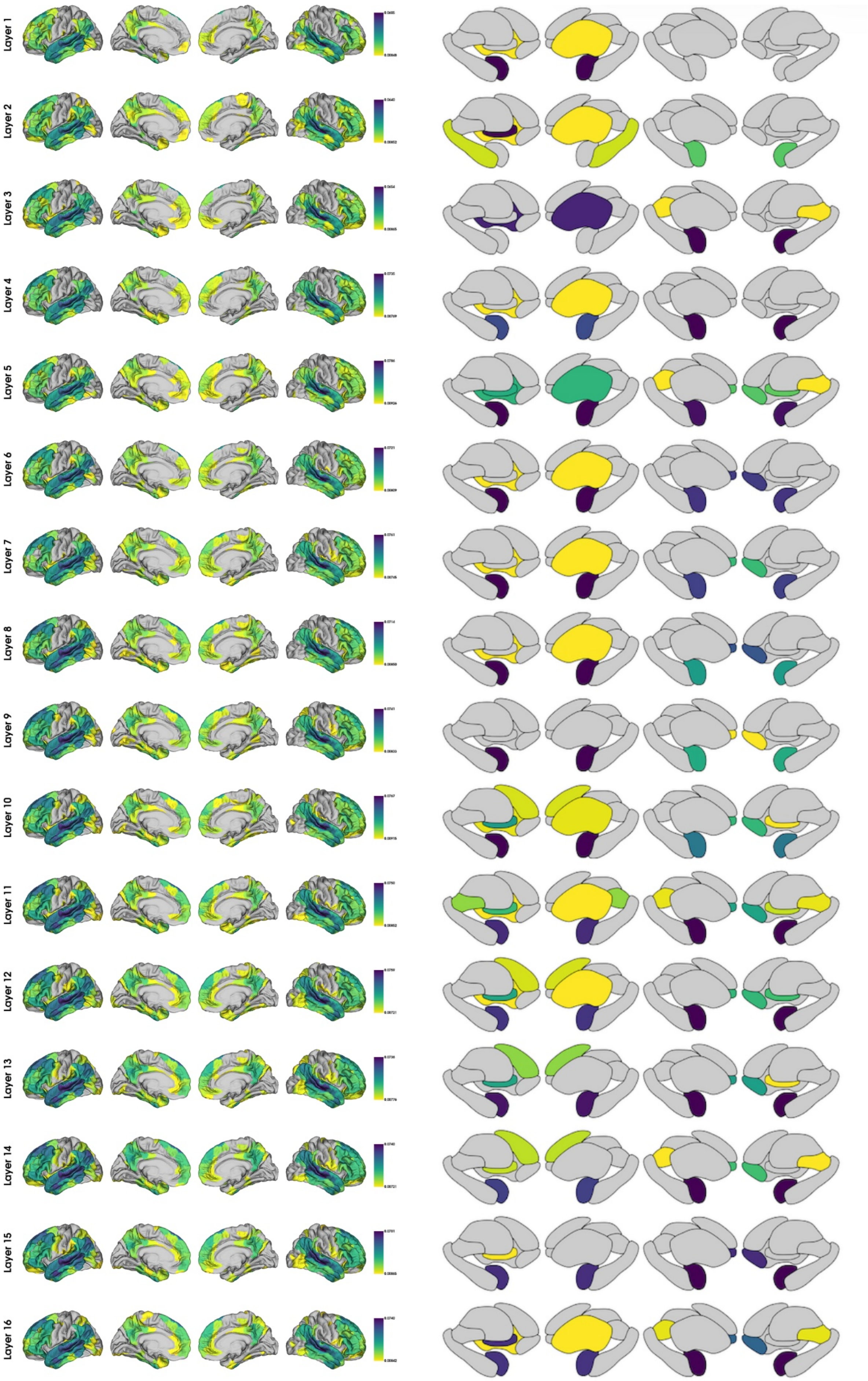
Layer 1
Layer 2
Layer 3
Layer 4
Layer 5
Layer 6
Layer 7
Layer 8
Layer 9
Layer 10
Layer 11
Layer 12
Layer 13
Layer 14
Layer 15
Layer 16

**Figure S19:** Cortical and subcortical prediction effects across layer 1-16 for Mistral English. Only regions with significant brain scores are colored. Statistical significance was assessed using one-sample t-tests of subject-level prediction scores against zero (FDR-corrected, $p < 0.05$).

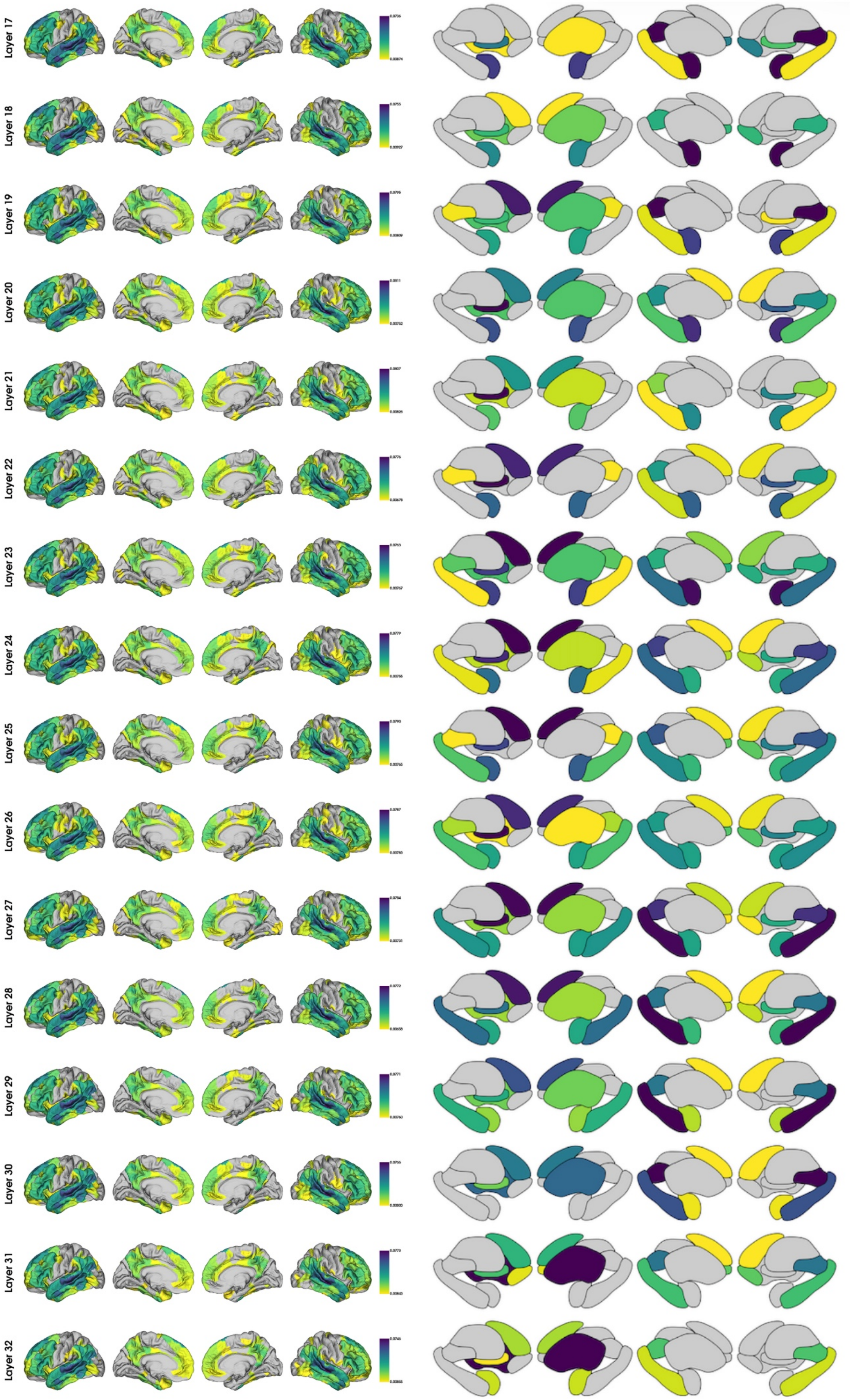
Layer 17
Layer 18
Layer 19
Layer 20
Layer 21
Layer 22
Layer 23
Layer 24
Layer 25
Layer 26
Layer 27
Layer 28
Layer 29
Layer 30
Layer 31
Layer 32

**Figure S20:** Cortical and subcortical prediction effects across layer 16-32 for Mistral English. Statistical significance was assessed using one-sample t-tests of subject-level prediction scores against zero (FDR-corrected, $p < 0.05$).

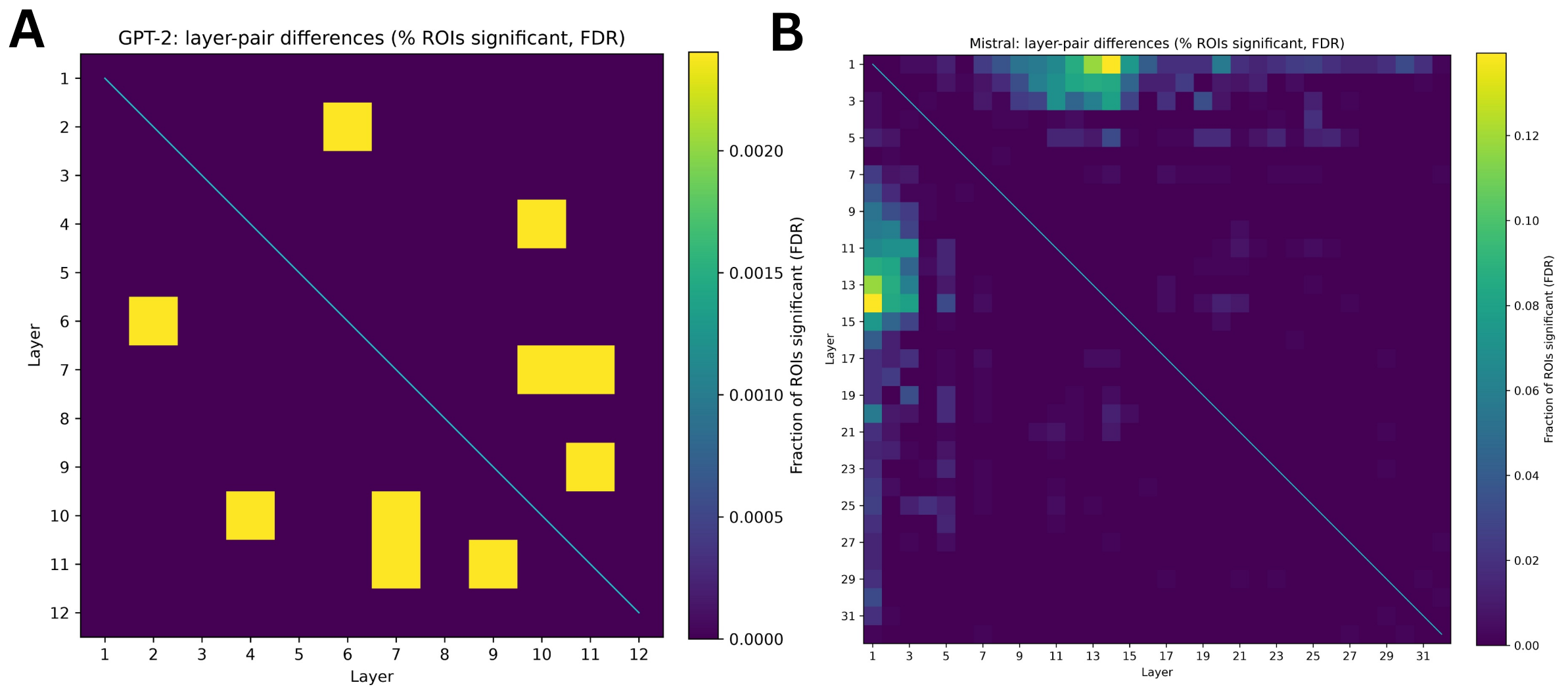

**Figure S21:** Pairwise differences across the layers of autoregressive language models. Cell colors indicate the proportion of ROIs showing a significant difference in prediction accuracy for each pair of layers; darker colors indicate few or no significantly different ROIs, whereas lighter colors indicate a larger proportion. The matrices are symmetric, and the diagonal denotes comparisons of each layer with itself. (A) Pairwise comparisons among GPT-2 Transformer layers 1-12. (B) Pairwise comparisons among Mistral Transformer layers 1-32. FDR correction was applied across ROIs within each layer pair and not additionally across layer pairs.

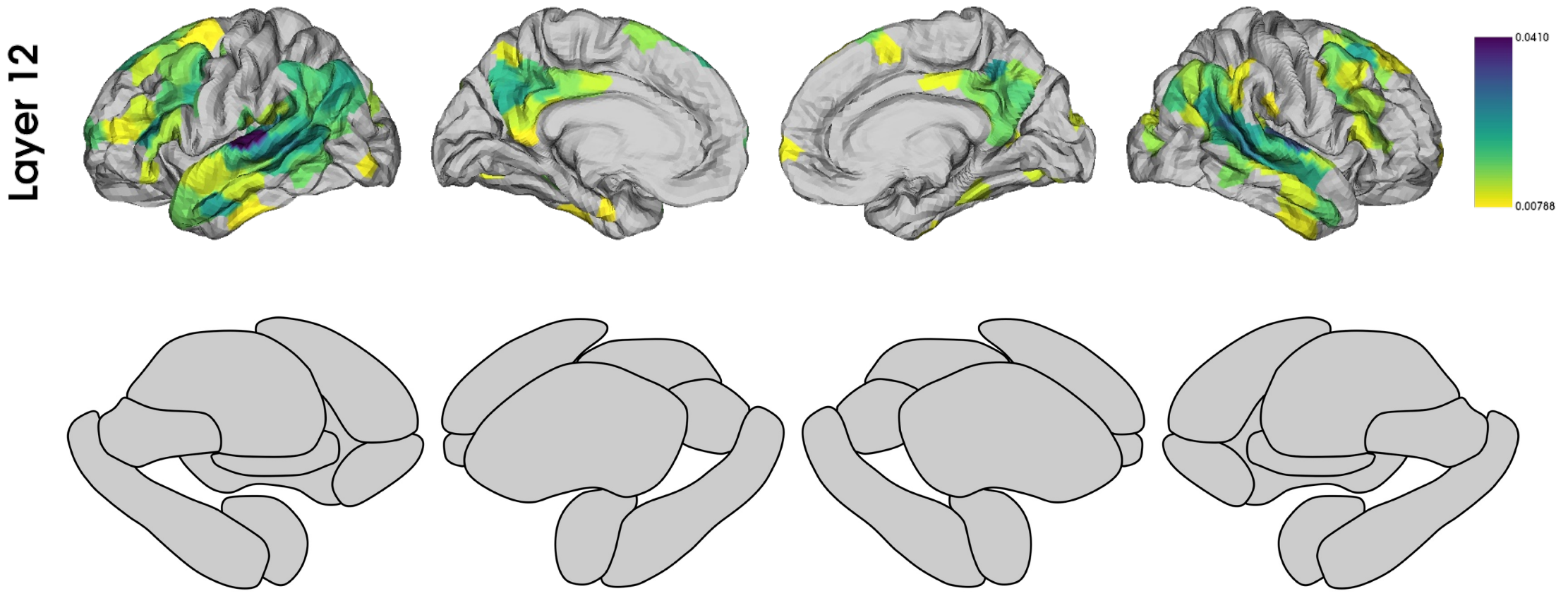

**Figure S22:** Significant predictions from Layer 12 of english mbert using FIR. Statistical significance was assessed using one-sample t-tests of subject-level prediction scores against zero (FDR-corrected, $p < 0.05$).

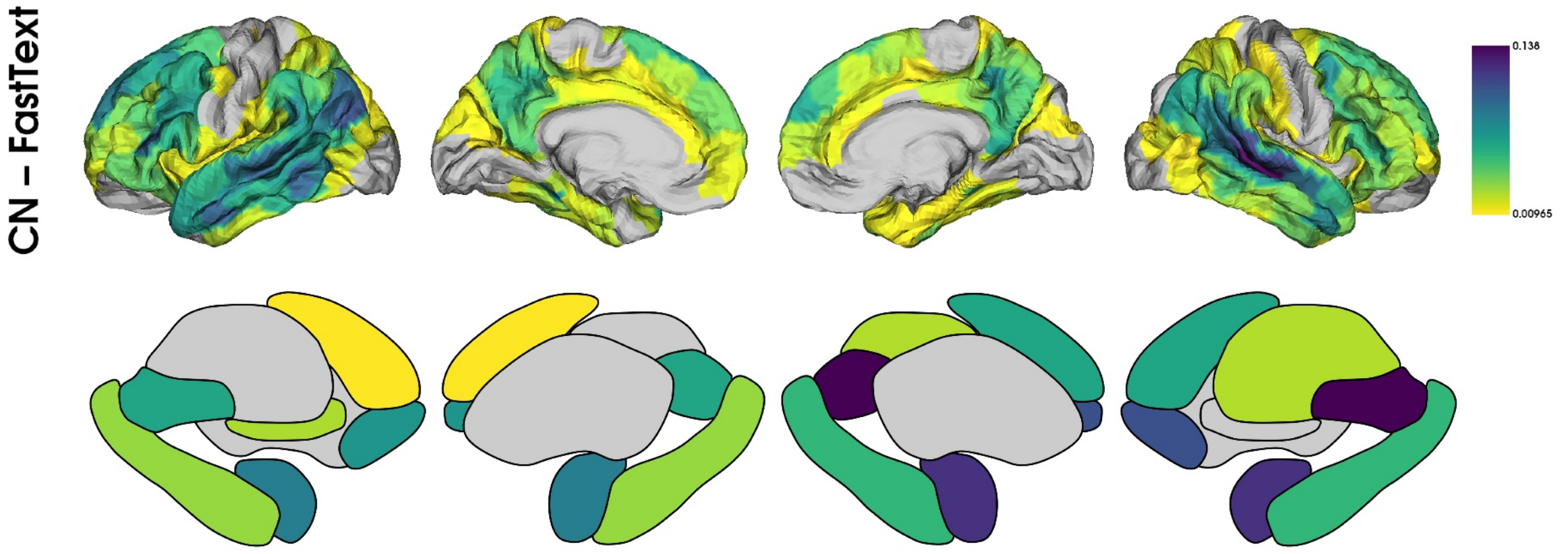

**Figure S23:** Significant predictions from Chinese FastText. Statistical significance was assessed using one-sample t-tests of subject-level prediction scores against zero (FDR-corrected, $p < 0.05$).

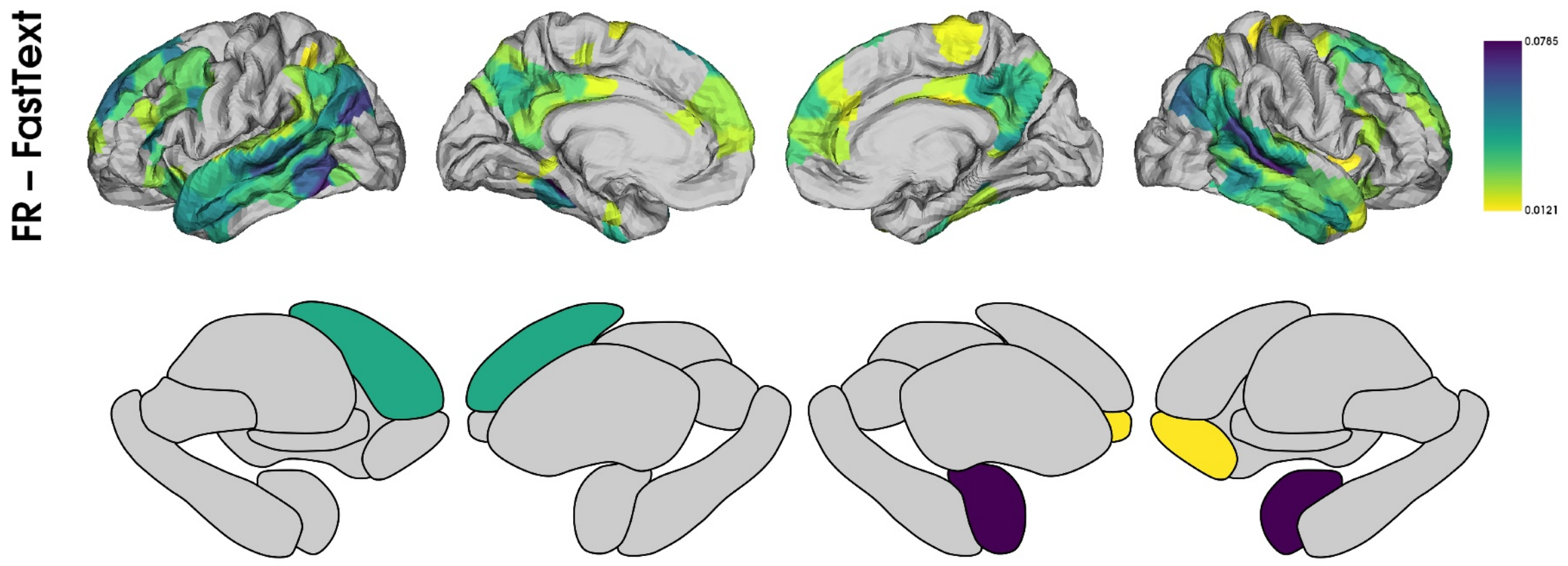

**Figure S24:** Significant predictions from French FastText. Statistical significance was assessed using one-sample t-tests of subject-level prediction scores against zero (FDR-corrected, $p < 0.05$).

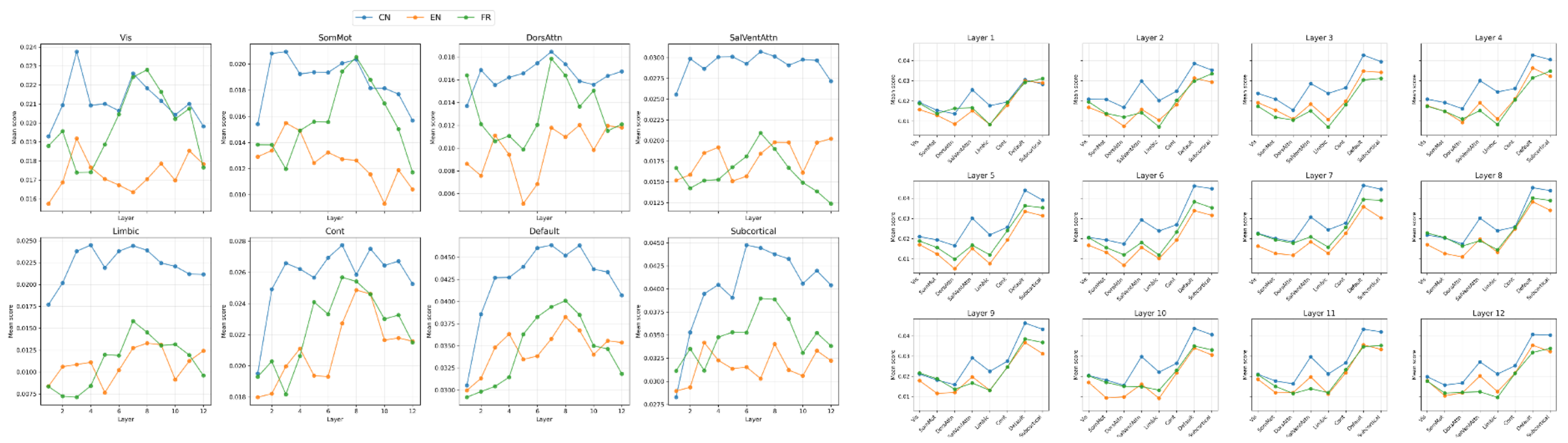

**Figure S25: :** Network-wise and layer-wise brain scores across languages for monolingual BERTs. Mean ROI-level brain scores within Yeo-7 cortical networks: visual (Vis), somatomotor (SomMot), dorsal attention (DorsAttn), ventral attention (SalVentAttn), limbic (Limbic), control (Cont), default mode (Default) network, and the Melbourne subcortical system in 12 mBERT layers for Chinese (CN), English (EN), and French (FR).

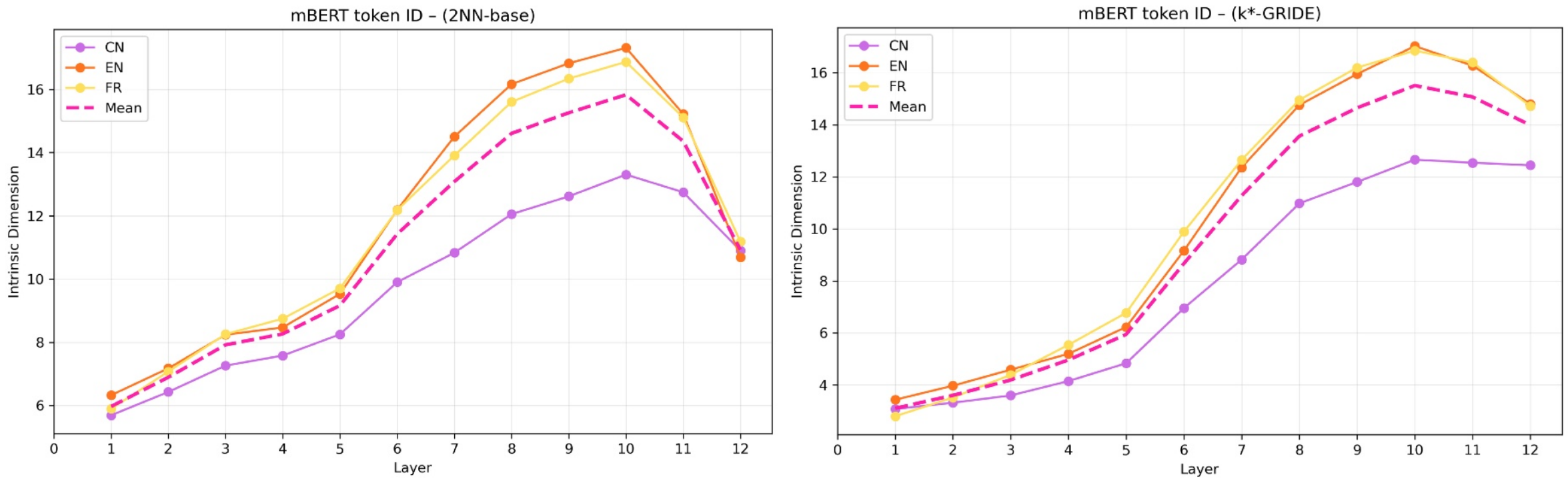

**Figure S26:** Layerwise intrinsic dimensionality estimated using 2NN-base and k*GRIDE methods.

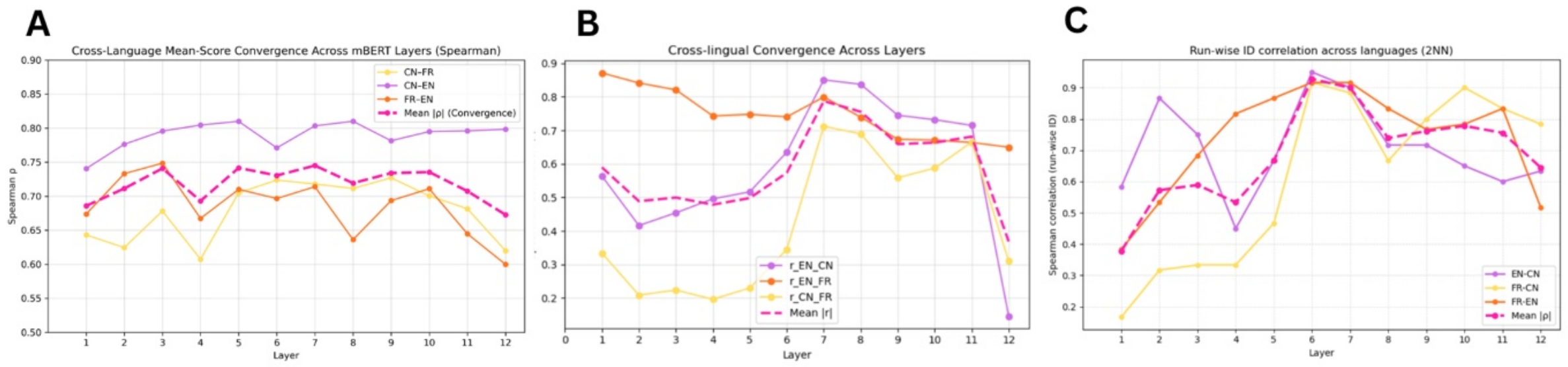

**Figure S27:** Cross-linguistic convergence. (A) To quantify cross-linguistic similarity of significant brain scores across mBERT layers, we computed layer-wise correlations between thresholded ROI-level score maps from each language. For each model layer, Spearman rank correlations were calculated across ROIs for each pair of languages (Chinese–French, Chinese–English, and French–English), using only ROIs with finite values in both languages. Correlations were treated as missing when fewer than two overlapping ROIs were available. As a summary measure of cross-linguistic convergence at each layer, we averaged the absolute correlations across the three language pairs, yielding a layer-wise profile of how similarly different languages' embeddings align with brain responses as model depth increases. (B) For surprisal convergence, we derived run-wise, layer-wise surprisal curves separately for each language using mBERT PLL surprisal. Because translations differ in word order, sentence structure, and the number of sentences across languages, analyses were restricted to runs present in all languages. For each sentence within each fMRI run, word-level PLL surprisal was computed for every transformer layer and then averaged within each run to obtain a run-level summary for each layer. Then at each layer, Pearson correlations were computed across runs for each pair of languages, using the same convergence procedure as for brain scores. An overall measure of cross-linguistic convergence at each layer was obtained by taking the absolute value of the mean pairwise correlation across the three language pairs. (C) ID was estimated for each run and layer-specific embedding set using the 2NN estimator as earlier.